\documentclass[10pt,twocolumn,letterpaper]{article}

\usepackage{iccv}
\usepackage{times}
\usepackage{epsfig}
\usepackage{graphicx}
\usepackage{amsmath}
\usepackage{amssymb}
\usepackage{booktabs}
\usepackage{threeparttable}
\usepackage{bbm}
\usepackage{float}
\usepackage{bbding}
\usepackage{subfigure}

\usepackage{marvosym}
\usepackage{pifont}
\newcommand{\cmark}{\ding{51}}%
\newcommand{\xmark}{\ding{55}}%

\usepackage[accsupp]{axessibility}  
\usepackage[pagebackref=true,breaklinks=true,letterpaper=true,colorlinks,bookmarks=false]{hyperref}
\newcommand{\tabincell}[2]{\begin{tabular}
{@{}#1@{}}#2\end{tabular}}

\newcommand\blfootnote[1]{%
  \begingroup
  \renewcommand\thefootnote{}\footnote{#1}%
  \addtocounter{footnote}{-1}%
  \endgroup
}

\iccvfinalcopy 

\def\iccvPaperID{6334} 

\ificcvfinal\fi

\begin{document}

\title{Relaxed Transformer Decoders for Direct Action Proposal Generation}

\author{Jing Tan\textsuperscript{*} \qquad Jiaqi Tang\textsuperscript{*} \qquad Limin Wang\textsuperscript{\Letter} \qquad Gangshan Wu \\
State Key Laboratory for Novel Software Technology, Nanjing University, China\\
{\tt\small \{jtan,jqtang\}@smail.nju.edu.cn, \{lmwang, gswu\}@nju.edu.cn}\\
}

\maketitle
\ificcvfinal\thispagestyle{empty}\fi

\begin{abstract}
   Temporal action proposal generation is an important and challenging task in video understanding, which aims at detecting all temporal segments containing action instances of interest. The existing proposal generation approaches are generally based on pre-defined anchor windows or heuristic bottom-up boundary matching strategies. This paper presents a simple and efficient framework (RTD-Net) for direct action proposal generation, by re-purposing a Transformer-alike architecture. To tackle the essential visual difference between time and space, we make three important improvements over the original transformer detection framework (DETR). First, to deal with slowness prior in videos, we replace the original Transformer encoder with a boundary attentive module to better capture long-range temporal information. Second, due to the ambiguous temporal boundary and relatively sparse annotations, we present a relaxed matching scheme to relieve the strict criteria of single assignment to each groundtruth. Finally, we devise a three-branch head to further improve the proposal confidence estimation by explicitly predicting its completeness.  Extensive experiments on THUMOS14 and ActivityNet-1.3 benchmarks demonstrate the effectiveness of RTD-Net, on both tasks of temporal action proposal generation and temporal action detection. Moreover, due to its simplicity in design, our framework is more efficient than previous proposal generation methods, without non-maximum suppression post-processing. The code and models are made available at \url{https://github.com/MCG-NJU/RTD-Action}.
\end{abstract}
\blfootnote{\small *: Equal contribution. \Letter: Corresponding author.}
\section{Introduction}

As large numbers of videos are captured and uploaded online (e.g., YouTube, Instagram, and TikTok), video understanding is becoming an important problem in computer vision. Action recognition~\cite{DBLP:conf/nips/SimonyanZ14,DBLP:conf/eccv/WangXW0LTG16,DBLP:conf/cvpr/CarreiraZ17,DBLP:conf/cvpr/WangL0G18,DBLP:journals/corr/abs-2104-09952,DBLP:journals/corr/abs-2012-10071} has received much research attention from both academia and industry, with a focus on classifying trimmed video clip into action labels. However, these action recognition methods cannot be directly applied for realistic video analysis due to the fact that these web videos are untrimmed in nature. Therefore, temporal action detection~\cite{DBLP:conf/eccv/LinZSWY18,DBLP:conf/iccv/LinLLDW19,DBLP:conf/iccv/ZengHGTRZH19,DBLP:conf/cvpr/XuZRTG20} is a demanding technique, which aims to localize each action instance in long untrimmed videos with the action category and as well its temporal duration.  In general, temporal action detection task is composed of two subtasks: temporal action proposal generation and action classification.

\begin{figure}[t]
\begin{center}
\includegraphics[scale=0.6]{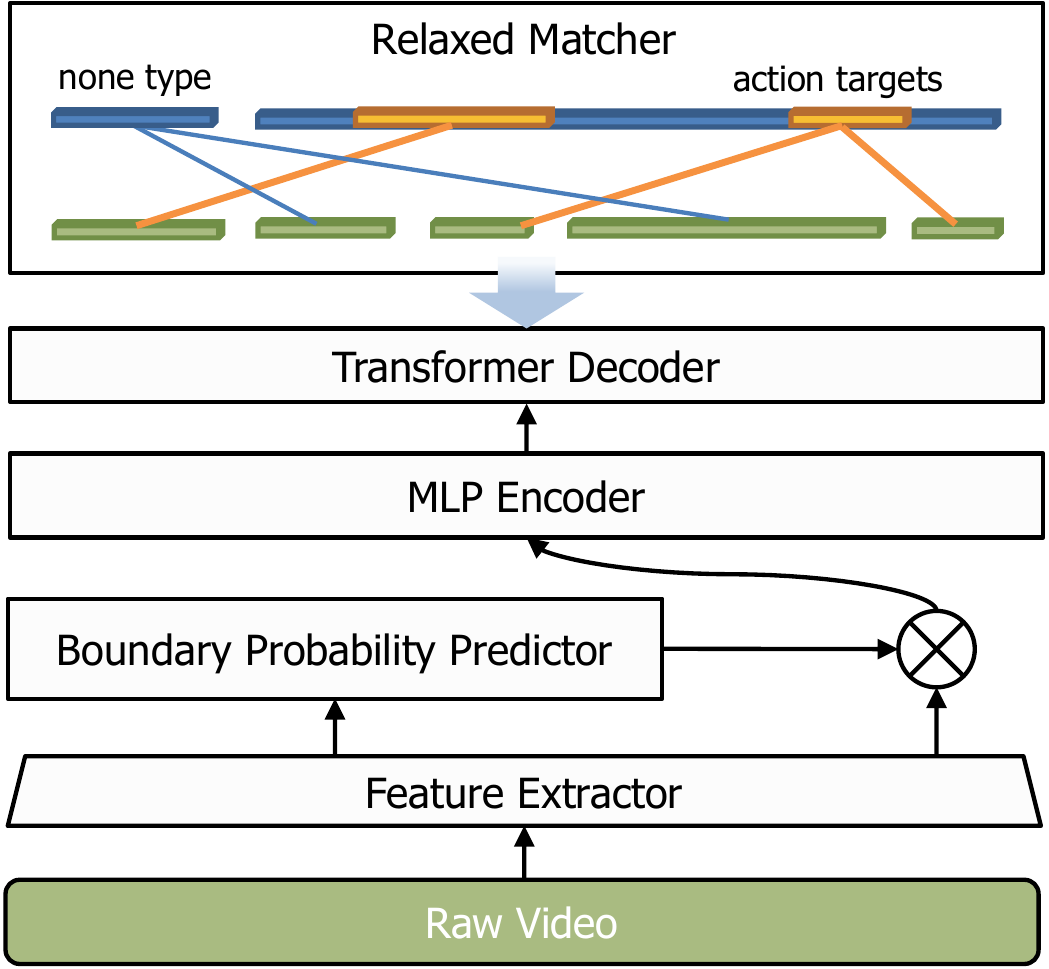}
\vspace{3mm}
\caption{{\bf Overview of RTD-Net}. Given an untrimmed video, RTD-Net directly generates action proposals based on boundary-attentive features without hand-crafted design, such as dense anchor placement, heuristic matching strategy, and non-maximum suppression.}
\label{fig:overview}
\end{center}
\vspace{-6mm}
\end{figure}

As for temporal proposal generation task, there are two mainstream approaches. The first type is an anchor-based~\cite{DBLP:conf/bmvc/BuchEGFN17, DBLP:conf/cvpr/HeilbronNG16, DBLP:conf/eccv/EscorciaHNG16, DBLP:conf/iccv/GaoYSCN17} method, which generates action proposals based on dense and multi-scale box placement. As the duration of action instances varies from seconds to minutes, it is almost impossible for these anchor-based methods to cover all these ground-truth instances under a reasonable computation consumption. The second type is a boundary-based~\cite{DBLP:conf/iccv/ZhaoXWWTL17, DBLP:conf/eccv/LinZSWY18, DBLP:conf/iccv/LinLLDW19} method, which first predicts the boundary confidence at all frames, and then employs a bottom-up grouping strategy to match pairs of start and end. These methods extract the boundary information at a local window and simply utilize the local context for modeling. Therefore, these boundary-based methods might be sensitive to noise and fail to yield robust detection results, as they easily produce incomplete proposals. Furthermore, the performance of these two kinds of methods is highly dependent on the carefully-designed anchor placement or sophisticated boundary matching mechanisms, which are hand-crafted with human prior knowledge and require specific tuning.

We contend that long-range temporal context modeling is vital for proposal generation. Viewing videos as temporal sequences and employing Transformer architecture to model global one-dimensional dependencies boosts localization performance. We propose a direct action proposal generation framework with Transformers. This direct action proposal generation with parallel decoding allows us to better capture inter-proposal relationships from a global view, thus resulting in more complete and precise localization results. Moreover, our temporal detection framework streamlines the complex action proposal generation pipeline with a neat set prediction paradigm, where hand-crafted designs such as anchor box placement, boundary matching strategy, and time-consuming non-maximum suppression are removed. As a result, our framework conducts inference with a noticeable faster speed. However, {\em due to the essential visual property difference between time and space, it is non-trivial to adapt the image detection Transformer architecture for videos}.

We observe that the feature slowness in videos~\cite{DBLP:journals/pami/ZhangT12} and ambiguous temporal boundaries~\cite{DBLP:conf/eccv/SatkinH10} are two key issues that require specific consideration for building a direct action proposal generation method with Transformers. First, although there are many frames along the temporal dimension, their features change at a very low speed.
Direct employment of self-attention mechanism as in Transformer encoder will lead to an over-smoothing issue and reduce the discrimination ability of action boundary. Second, due to the high-level semantic for action concept, its temporal boundary might be not so clear as object boundary, and the ground-truth labels might also contain some noise due to inconsistency among different labors. So a strict set matching loss might have a negative effect on the convergence of Transformer, and not be optimal for training and generalization.

To address the above issues, we present a Relaxed Transformer Decoder (RTD) architecture for direct action proposal generation, as shown in Figure~\ref{fig:overview}. Compared with the original object detection Transformer, we make three notable improvements to adapt for the video task. \emph{First}, we replace the original Transformer encoder with a customized boundary-attentive architecture to overcome the over-smoothing issue.
\emph{Second}, we propose a relaxed matcher to relieve the strict criteria of single assignment to a ground-truth. 
\emph{Finally}, we devise a three-branch detection head for training and inference. A completeness head is added to explicitly estimate the tIoU between regressed temporal box and ground-truth box. We observe that this tIoU loss can guide the training of Transformer and regularize three heads to converge to a stable solution. 

In summary, our main contributions are as follows:
\begin{itemize}
    \item For the first time, we adapt the Transformer architecture for direct action proposal generation in videos to model inter-proposal dependencies from a global view, and reduce the inference time greatly by streamlining temporal action proposal generation pipeline with a simple and neat framework, removing the hand-crafted designs.
    \item We make three important improvements over DETR~\cite{DBLP:conf/eccv/CarionMSUKZ20} to address the essential difference between temporal location in videos and spatial detection in images, including boundary attentive representation, relaxation mechanism, and three-branch head design. 
    \item Experiments demonstrate that our method outperforms the existing state-of-the-art methods on THUMOS14 and achieves comparable performance on  ActivityNet-1.3, in both temporal action proposal generation task and temporal action detection task.
\end{itemize}

\begin{figure*}[t]
\begin{center}
\includegraphics[scale=0.45]{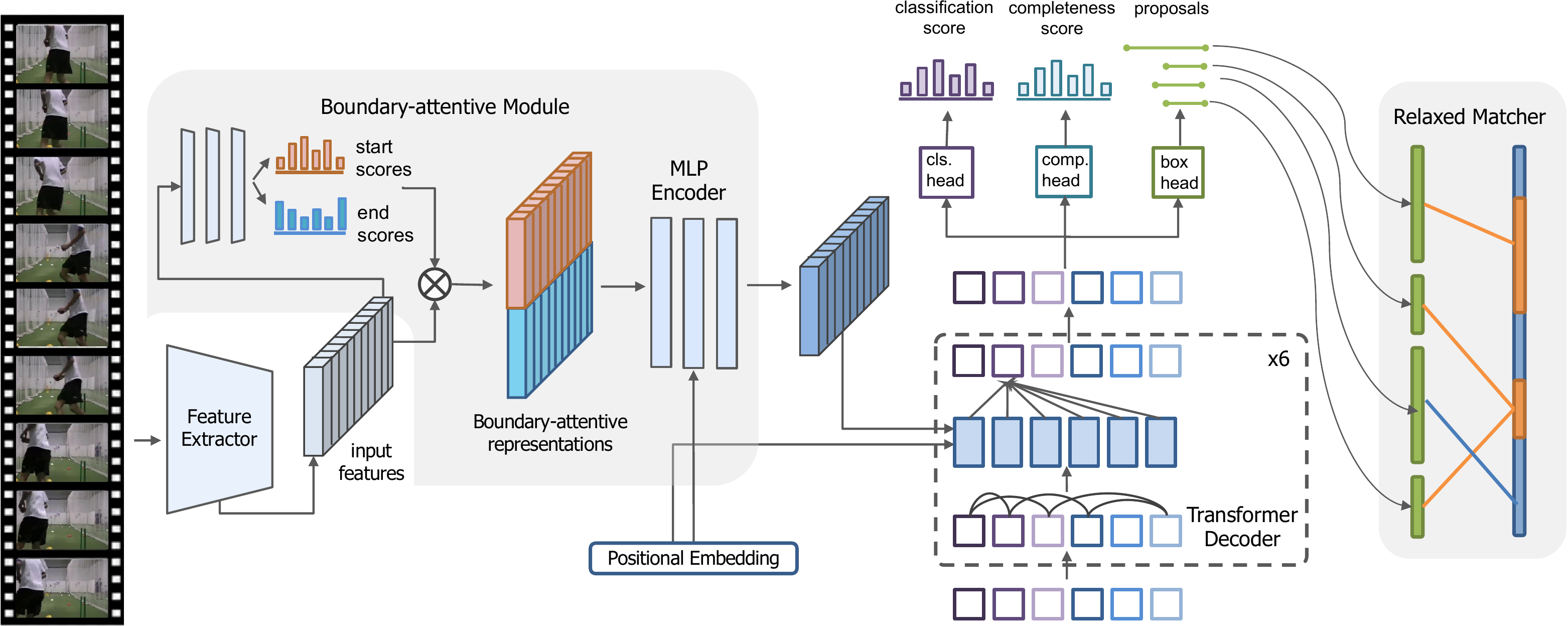}
\end{center}
   \caption{{\bf Pipeline of RTD-Net.} Our RTD-Net streamlines the process of temporal action proposal generation by viewing it as a direct set prediction problem. It is composed of three unique designs: a boundary-attentive module for feature extraction, a transformer decoder for direct and parallel decoding of queries, and a relaxed matcher for training label assignment. Our RTD-Net is able to efficiently generate a set of smaller number of proposals without any post processing. }
\label{fig:long}
\label{Fig_Arch}
\end{figure*}

\section{Related Work}
\noindent\textbf{Action Recognition.} Action recognition is a fundamental task in video understanding, the same as image classification in the image domain. In addition to provide semantic labels for trimmed videos, action recognition is also eligible for extracting snippet-level features in untrimmed videos, which are used in downstream tasks, such as temporal action detection~\cite{DBLP:conf/iccv/ZhaoXWWTL17,DBLP:conf/iccv/LinLLDW19}, language-based video grounding~\cite{DBLP:conf/aaai/ZhangPFL20, DBLP:conf/cvpr/ZengXHCTG20}, and spatio-temporal action detection~\cite{DBLP:conf/eccv/LiW0W20, DBLP:conf/iccv/KalogeitonWFS17a}. There are two main types of video architectures: two-stream networks~\cite{DBLP:conf/nips/SimonyanZ14, DBLP:conf/eccv/WangXW0LTG16, DBLP:conf/cvpr/FeichtenhoferPZ16} extracted video appearance and motion information from RGB image and stacked optical flow; 3D convolution networks~\cite{DBLP:conf/iccv/TranBFTP15, DBLP:conf/cvpr/CarreiraZ17, DBLP:conf/iccv/QiuYM17} directly captured appearance and motion clues with spatio-temporal kernels. We use I3D~\cite{DBLP:conf/cvpr/CarreiraZ17} model to extract video feature sequences as RTD-Net input.

\noindent\textbf{Temporal Action Proposal Generation.} The goal of temporal action proposal generation is to generate proposals in untrimmed videos flexibly and precisely. Among temporal action proposal generation methods, anchor-based methods~\cite{DBLP:conf/bmvc/BuchEGFN17, DBLP:conf/cvpr/HeilbronNG16, DBLP:conf/eccv/EscorciaHNG16, DBLP:conf/iccv/GaoYSCN17, DBLP:conf/iccv/XuDS17, DBLP:conf/cvpr/ChaoVSRDS18} retrieved proposals based on multi-scale and dense anchors, which is inflexible and cannot cover all action instances. Boundary-based methods~\cite{DBLP:conf/iccv/ZhaoXWWTL17, DBLP:conf/eccv/LinZSWY18, DBLP:conf/iccv/LinLLDW19, DBLP:conf/aaai/LinLWTLCWLHJ20} first evaluated the confidence of starting and ending points and then matched them to form proposal candidates. However, they generated results only based on local information and were easily affected by noise. On the contrary, our framework makes predictions based on the whole feature sequence and fully leverages the global temporal context. Recently, graph-based methods~\cite{DBLP:conf/cvpr/XuZRTG20, DBLP:conf/eccv/BaiWTYLL20} gained popularity in this field, they exploited long-range context based on pre-defined graph structure, the construction of which is highly dependent on human prior knowledge. In contrast, RTD-Net learns its own queries and directly generates complete and precise proposals without any hand-crafted design (anchor matching strategy or graph construction), and is free of time-consuming NMS module.

\noindent\textbf{Transformer and Self-Attention Mechanism.} 
Transformer was firstly introduced by~\cite{DBLP:conf/nips/VaswaniSPUJGKP17} in machine translation task. It tackles the problem of long-range dependency modeling in sequence modeling task. The major feature in Transformer is the self-attention mechanism, which summarizes content from the source sequence and is capable of modeling complex and arbitrary dependencies within a limited number of layers. Inspired by the recent advances in NLP tasks~\cite{DBLP:conf/naacl/DevlinCLT19, DBLP:conf/acl/DaiYYCLS19}, self-attention was applied to vision tasks to leverage large-scale or long-range context. For instance, works based on self-attention blocks appeared in image generation~\cite{DBLP:conf/icml/ParmarVUKSKT18}, image recognition~\cite{DBLP:conf/icml/ParmarVUKSKT18, DBLP:conf/nips/ParmarRVBLS19, DBLP:conf/iclr/CordonnierLJ20}
, action recognition~\cite{DBLP:conf/cvpr/GirdharCDZ19} and object detection~\cite{DBLP:conf/eccv/CarionMSUKZ20}. Some~\cite{DBLP:conf/icml/ParmarVUKSKT18, DBLP:conf/nips/ParmarRVBLS19, DBLP:conf/iclr/CordonnierLJ20} used specialized self-attention blocks as substitutes for convolutions, and others used self-attention blocks to replace components in the convolution networks. Recent work~\cite{DBLP:journals/corr/abs-2010-11929}~showed that with Transformer architecture alone, self-attention blocks could achieve excellent results for image recognition. We use decoder-only Transformer on videos for temporal proposal generation, where our model is able to fully exploit the global temporal context and generate action proposals in a novel and direct paradigm.

\section{Method}

\subsection{Overview}

We propose a relaxed transformer decoder network (RTD-Net) to directly generate temporal action proposals. Specifically, given an input video $X$ with $l_f$ frames, RTD-Net aims at generating a set of proposals $\Psi = \{\psi_n = (t_s^n, t_e^n)\}$, locating the underlying human action instances $\hat{\Psi} = \{\hat{\psi}_n = (\hat{t}_s^n, \hat{t}_e^n)\}_{n=1}^{N_g}$, with $N_g$ as the number of action instances in video $X$.

In order to tackle the issues caused by feature slowness and ambiguous temporal boundary, RTD-Net features three major components: a boundary-attentive module, a relaxed Transformer decoder, and a three-branch detection head. The main architecture is illustrated in Figure \ref{Fig_Arch}. First, we use backbone network to extract short-term features. Then the boundary-attentive module enhances them with discriminative boundary scores, and outputs compact boundary-attentive representations to be fed into a transformer decoder. As shown in experiment, we find that this boundary-attentive module is important for the subsequent decoding process. After this, the transformer decoder uses a set of learned queries to attend to the boundary-attentive representations. This parallel decoding process is able to model all pair-wise constraints among proposal candidates explicitly and capture inter-proposal context information with a global view. Eventually, a three-branch detection head transforms the decoder embedding to our final prediction results. Boundary head directly generates temporal boxes, and binary classification head combined with completeness head gives a confidence score for each predicted box. For training, we give a relaxed matching criteria in the matcher, which alleviates the impact of ambiguous temporal boundaries and allows more well-predicted proposals to be assigned as positive samples.

\subsection{Feature Encoding}
We adopt two-stream networks~\cite{DBLP:conf/eccv/WangXW0LTG16,DBLP:conf/cvpr/CarreiraZ17} to extract appearance features $F_A = \{f_{t_n,A}\}$ from RGB frames and motion features $F_M = \{f_{t_n,M}\}$ from stacked optical flow at time $t_n$. Features are extracted with a sliding window of temporal stride $\tau$ and arranged into a sequence of length $l_s$. Following the common practice, we take features after the global pooling layer and before the top fully-connected layer from feature extractor networks. Appearance features and motion features are concatenated along channel dimension to form our final input feature sequence $F = \{f_{t_n}\}_{n=1}^{l_s}$, where $f_{t_n} = (f_{t_n,A} , f_{t_n, M}) $.

\subsection{Direct Action Proposal Generation Mechanism}

\noindent\textbf{Boundary-attentive representations.}
As analyzed above, slowness is a general prior for video data, where short-term features change very slowly in a local window. Meanwhile, our short-term features are usually extracted from a short video segment with overlap, which will further smooth visual features. For temporal action proposal generation, it is crucial to keep sharp boundary information in visual representations to allows for the subsequent decoding processing. To alleviate the issue of feature slowness, we propose the boundary-attentive module to explicitly enhance short-term features with discriminative action boundary information. Specifically, we multiply the original features with its own action starting and ending scores, where the scores of action boundary at each time are estimated with a temporal evaluation module~\cite{DBLP:conf/eccv/LinZSWY18}. In experiments, we find that this boundary-attentive representation is helpful for our transformer decoder to generate more accurate action proposals, thanks to the explicit leverage of action boundary information. MLP encoder is employed to transform the boundary-attentive representation into a more compact form.

\noindent\textbf{Relaxed Transformer decoder.}
We use the vanilla Transformer decoder to directly output temporal action proposals. The decoder takes a set of proposal queries and boundary-attentive representations as input, and outputs action proposal embedding for each query via stacked multi-head self-attention and encoder-decoder attention blocks. Self-attention layers model the temporal dependencies between proposals and refine the corresponding query embedding. In ‘encoder-decoder’ attention layers, proposal queries attend to all time steps and aggregate action information at high activation into each query embedding. During training procedure, this decoder collaborates with a Hungarian matcher to align positive proposals with groundtruth and the whole pipeline is trained with a set prediction loss.

Unlike object detection in images, temporal action proposal generation is more ambiguous and sparse in annotation. For instance, only a few actions appear in an observation window for THUMOS14 and the average number of action instances in ActivityNet-1.3 is only 1.5. In addition, the temporal variation of action instances is significant across different videos, in particular for ActivityNet dataset. So, the matching criteria that only a single detection result matches a groundtruth instance might be sub-optimal for temporal action proposal generation. In practice, we observe that the visual difference between some temporal segments around the groundtruth is very small, and the strict matching criteria will make whole network confused and thereby hard to converge to a stable solution.

To deal with this problem, we propose a relaxed matching scheme, where multiple detected action proposals are assigned as positive when matching to the groundtruth. Specifically, we use a tIoU threshold to distinguish positive and negative samples, where tIoU is calculated as the intersection between target and prediction over their union. The predictions with tIoU higher than a certain threshold will be marked as positive samples. In experiments, we observe that this simple relaxation will relieve the training difficulty of RTD-Net and is helpful to improve the final performance.

\noindent\textbf{Three-branch head design.}
RTD-Net generates final predictions by designing three feed forward networks (FFNs) as detection heads. We generalize the box head and class head in object detection to predict temporal action proposals. Boundary head decodes temporal boundary tuple of an action proposal $\psi_n = (t_s^n, t_e^n)$, which consists of a starting frame $t_s^n$ and an ending frame $t_e^n$. Binary classification head predicts foreground confidence score $p_{bc}$ of each proposal. In addition, a completeness head is proposed to evaluate prediction completeness $p_c$ with respect to the groundtruth.
 
A high-quality proposal requires not only high foreground confidence but also accurate boundaries. Sometimes, the binary classification score alone fails to be a reliable measure of predictions due to the confused action boundaries. RTD-Net introduces a completeness head to predict a completeness score $p_c$ that measures the overlap between predictions and targets. This additional completeness score is able to explicitly incorporate temporal localization quality to improve the proposal confidence score estimation, thus making the whole pipeline more stable.

\subsection{Training}

In our training, we first scale video features into a fixed length for subsequent processing. Specifically, following the common practice, we employ a sliding window strategy with a fixed overlap ratio on THUMOS14 dataset and a re-scaling operation on ActivityNet-1.3 dataset. In THUMOS14, only observation windows that contain at least one target are selected for training.

\noindent\textbf{Boundary-attentive module. }
Starting and ending scores are predicted as boundary probabilities. We follow the footsteps of BSN~\cite{DBLP:conf/eccv/LinZSWY18}, and use a three-layer convolution network as the boundary probability predictor. This predictor is trained in a frame level to generate starting and ending probability $p_{t_n,s}$ and $p_{t_n,e}$ for each temporal location $t_n$.

\noindent\textbf{Label assignment of RTD-Net.}
The ground-truth instance set  $\hat{\Psi} = \{\hat{\psi}_n = (\hat{t}_s^n, \hat{t}_e^n)\}_{n=1}^{N_g}$ is composed of $N_g$ targets, where $\hat{t}_{s}^n$ and $\hat{t}_{e}^n$ are starting and ending temporal locations of $\hat{\psi}_n$. Likewise, the prediction set of $N_p$ samples is denoted as $\Psi = \{
\psi_n = (t_s^n, t_e^n)\}_{n=1}^{N_p}$. We assume $N_p$ is larger than $N_g$ and augment $\hat{\Psi}$ to be size $N_p$ by padding $\emptyset$. Similar to DETR~\cite{DBLP:conf/eccv/CarionMSUKZ20}, RTD-Net first searches for an optimal bipartite matching between these two sets and the cost of the matcher is defined as:

\begin{small}
\begin{equation}
    C = \sum_{n:\sigma(n) \neq \emptyset} \alpha\,\cdot\,\ell_1(\psi_n, \hat{\psi}(\sigma(n)))-\beta~\cdot~tIoU(\psi_n,\hat{\psi}(\sigma(n))-\gamma \cdot {p_{bc,n}},
\end{equation}
\end{small}
where $\sigma$ is a permutation of $N_p$ elements to match the prediction to targets, $\alpha$, $\beta$, and $\gamma$ are hyper-parameters and specified as $1$, $5$, $2$ in experiments. Here we use both $\ell_1$ loss and $tIoU$ loss for bipartite matching due to its complementarity.
Based on the Hungarian algorithm, the matcher is able to search for the best permutation with the lowest cost. Besides, a relaxation mechanism is proposed to address sparse annotations and ambiguous boundaries of action instances. We calculate tIoU between targets and predictions, and also mark those predictions with tIoU higher than a certain threshold as positive samples. After relaxation, the updated assignment of predictions is notated as $\sigma'$.

\noindent\textbf{Loss of the binary classification head.}
We define the binary classification loss function as:
\begin{equation}
    L_{cls} = -\gamma~\cdot~\frac{1}{N}\sum_{n=1}^{N}
    (\hat{p}_n\log p_{bc,n} + (1-\hat{p}_n)\log(1-p_{bc,n})),
\label{Class_loss}
\end{equation}
where $p_{bc,n}$ is the binary classification probability and $N$ is the total number of training proposals. $\hat{p}_n$ is 1 if the sample is marked positive, and otherwise 0.

\noindent\textbf{Loss of the boundary head.} Training loss function for the boundary head is defined as follows: 
\begin{equation}
    L_{boundary} = \frac{1}{N_{pos}}\sum_{n:\sigma'(n)\neq \emptyset} (\alpha~\cdot~L_{loc, n} + \beta\;\cdot\;L_{overlap, n}), 
\label{Boundary_loss}
\end{equation}
where $\ell_1$ loss is used in localization loss and tIoU loss is used in overlap measure:
\begin{equation}
\begin{gathered} 
L_{loc, n} = ||\hat{t}_s^{\sigma'(n)} - t_s(n)||_{l1} + || \hat{t}_e^{\sigma'(n)} - t_e(n)||_{l1},\\
L_{overlap, n} = 1 - tIoU(\psi_n, \hat{\psi}(\sigma'(n))).
\end{gathered}
\end{equation}

\noindent\textbf{Loss of the completeness head.}
To generate a robust and reliable measure of predictions, we introduce a completeness head to aid the binary classification head. Each proposal sampled for training calculates the tIoU with all targets, and the maximum tIoU is denoted as $\hat{g}_{tIoU}$. We adopt temporal convolution layers followed with one fully connected layer upon decoder outputs to predict completeness. To guide the training completeness branch, a loss based on tIoU is proposed:
\vspace{-2mm}
\begin{equation}
    L_{complete} = \frac{1}{N_{train}}\sum_{n=1}^{N_{train}}(p_{c, n} - \hat{g}_{tIoU, n})^2.
\end{equation}
At the beginning, the boundary head fails to predict high-quality proposals, and thus the completeness head cannot be effectively trained with low-quality proposals. We follow DRN~\cite{DBLP:conf/cvpr/ZengXHCTG20} to apply a two-step training strategy. In the first step, we freeze the parameters of the completeness head and train RTD-Net by minimizing Equations (\ref{Class_loss}) and (\ref{Boundary_loss}). In the second step, we fix other parts of RTD-Net and only train the completeness head.
\vspace{-2mm}
\subsection{Inference}
Due to the direct proposal generation scheme in our RTD, we follow a simple proposal generation pipeline without post-processing methods as non-maximum suppression, that are widely used in previous methods \cite{DBLP:conf/iccv/LinLLDW19, DBLP:conf/eccv/LinZSWY18, DBLP:conf/cvpr/ChaoVSRDS18}.

\noindent\textbf{Boundary-attentive module. }
To preserve the magnitude of features, we normalize the probability sequence $\{(p_{t_n, s}, p_{t_n, e})\}_{n=1}^{l_s}$ to the range of [0,1] and then scale it by $\alpha_r$. $\alpha_r$ is a scaling factor that re-scales boundary scores for stronger discrimination ability, and its choice will be discussed in ablation study. Feature sequence $F = \{f_{t_n}\}_{n=1}^{l_s}$ are multiplied with starting and ending scores separately, and then concatenated along channel dimension. Equipped with positional embedding, the visual representations are forward to a three-layer MLP encoder. Positional embedding is introduced here for temporal discrimination. The MLP encoder models the channel correlation and compacts boundary-attentive representations. 

\noindent\textbf{Proposal generation.} In Transformer decoder, the previous boundary attentive representations are directly retrieved with a set of learnt queries. In the end, for each query, three heads directly output its proposal boundaries, binary classification score, and completeness score.

\noindent\textbf{Score fusion.}
To make a more reliable confidence estimation for each proposal, we fuse the binary classification score $p_{bc}$ and completeness score $p_{c}$ for each proposal with a simple average. The resulted final proposal set is directly evaluated without any post-processing method.

\section{Experiments}

\subsection{Dataset and Setup}
\noindent\textbf{THUMOS14 \cite{THUMOS14}.} THUMOS14 dataset consists of 1,010 validation videos and 1,574 testing videos of 101 action classes in total. Among them 20 action classes are selected for temporal action detection. It contains 200 and 213 untrimmed videos with temporal annotations in validation and testing sets.

\noindent\textbf{ActivityNet-1.3~\cite{DBLP:conf/cvpr/HeilbronEGN15}.} ActivityNet-1.3 dataset contains 19,994 untrimmed videos with 200 action categories temporally annotated, and it is divided into training, validation and testing sets by the ratio of 2:1:1.

\noindent\textbf{Implementation details.} We adopt two-stream network TSN~\cite{DBLP:conf/eccv/WangXW0LTG16} and I3D~\cite{DBLP:conf/cvpr/CarreiraZ17} for feature encoding. Since TSN features better preserve local information, they are fed into the temporal evaluation module~\cite{DBLP:conf/eccv/LinZSWY18} for boundary confidence prediction. Compared with TSN features, I3D features have larger receptive fields and contain more contextual information. I3D features are enhanced by boundary probabilities and then input into MLP encoder for transformation and compression. During THUMOS14 feature extraction, the frame stride is set to 8 and 5 for I3D and TSN respectively. As for ActivityNet-1.3, the sampling frame stride is 16.  

On THUMOS14, we perform proposal generation in a sliding window manner and the length of each sliding window is set to 100 and the overlap ratio is set to 0.75 and 0.5 at training and testing respectively. As for ActivityNet-1.3,  feature sequences are rescaled to 100 via linear interpolation. To train RTD-Net from scratch, we use AdamW for optimization. The batch size is set to 32 and the learning rate is set to 0.0001.

\begin{table}
\begin{center}
\caption{Comparison with other state-of-the-art proposal generation methods on the test set of THUMOS14 in terms of AR@AN. SNMS stands for Soft-NMS.}
\vspace{1mm}
\scalebox{0.85}{
\begin{tabular}{cccccc}
\toprule[1pt]
Method & @50 & @100 & @200 & @500  \\ 
\hline
TAG+NMS~\cite{DBLP:conf/iccv/ZhaoXWWTL17} & 18.55 & 29.00 & 39.61 & -  \\ 
TURN+NMS~\cite{DBLP:conf/iccv/GaoYSCN17}  & 21.86 & 31.89 & 43.02 & 57.63  \\ 
CTAP+NMS~\cite{DBLP:conf/eccv/GaoCN18} & 32.49 & 42.61 & 51.97 & -  \\ 
BSN+SNMS~\cite{DBLP:conf/eccv/LinZSWY18} & 37.46 & 46.06 & 53.21 & 60.64  \\ 
BSN*+SNMS & 36.73 & 44.14 & 49.12 & 52.26  \\
MGG~\cite{DBLP:conf/cvpr/Liu0Z0C19} & 39.93 & 47.75 & 54.65 & 61.36 \\
BMN+SNMS~\cite{DBLP:conf/iccv/LinLLDW19} & 39.36 & 47.72 & 54.70 & 62.07 \\ 
BMN*+SNMS & 37.03 & 44.12 & 49.49 & 54.27  \\
DBG+SNMS~\cite{DBLP:conf/aaai/LinLWTLCWLHJ20} & 37.32 & 46.67 & 54.50 & 62.21  \\
RapNet+SNMS~\cite{DBLP:conf/aaai/GaoSWLYGZ20} & 40.35 & 48.23 & 54.92 & 61.41 \\
BC-GNN+SNMS~\cite{DBLP:conf/eccv/BaiWTYLL20} & 40.50 & \textbf{49.60} & 56.33 & 62.80  \\
\hline
RTD-Net* & \textbf{41.52} & {49.32} & \textbf{56.41} & \textbf{62.91}  \\
\bottomrule[1pt]
\end{tabular}
}
\begin{tablenotes}
\footnotesize
\item * results are reported based on P-GCN I3D features.
\end{tablenotes}
\label{thumos_prop}
\vspace{-8mm}
\end{center}
\end{table}

\begin{table}
\begin{center}
\caption{Comparison with other state-of-the-art proposal generation methods on validation set of ActivityNet-1.3 in terms of AR@AN and AUC. Among them, only RTD-Net is free of NMS.}
\vspace{1mm}
\scalebox{0.60}{
\begin{tabular}{ccccccc}
\toprule[1pt]
Method & \cite{DBLP:journals/corr/LinZS17} & CTAP~\cite{DBLP:conf/eccv/GaoCN18} & BSN~\cite{DBLP:conf/eccv/LinZSWY18} & MGG~\cite{DBLP:conf/cvpr/Liu0Z0C19} & BMN~\cite{DBLP:conf/iccv/LinLLDW19} & RTD-Net \\ 
\hline
AR@1 (val) & - & - & 32.17 & - & - & \textbf{33.05} \\ 
AR@100 (val) & 73.01 & 73.17 & 74.16 & 74.54 & \textbf{75.01} & 73.21 \\ 
AUC (val) & 64.40 & 65.72 & 66.17 & 66.43 & \textbf{67.10} & 65.78 \\
\bottomrule[1pt]
\end{tabular}
}
\label{ANet}
\vspace{-7mm}
\end{center}
\end{table}

\subsection{Temporal Action Proposal Generation}
\noindent\textbf{Evaluation metrics.} To evaluate the quality of proposals, we calculate Average Recall (AR) with Average Number (AN) of proposals and area under AR vs AN curve per video, which are denoted by AR@AN and AUC. Following the standard protocol, we use tIoU thresholds set [0.5 : 0.05 : 1.0] on THUMOS14 and [0.5 : 0.05 : 0.95] on ActivityNet-1.3. 

\noindent\textbf{Comparison with state-of-the-art methods.} Due to the high discriminative power of I3D features, we use it in our RTD-Net for proposal generation. For fair comparison, we also implement BSN~\cite{DBLP:conf/eccv/LinZSWY18} and BMN~\cite{DBLP:conf/iccv/LinLLDW19} with the same I3D features by the public available code. The experiment results on THUMOS14 are summarized in Table~\ref{thumos_prop}. Since BSN and BMN are highly dependent on the local context, therefore its performance drops on I3D features with large receptive fields. The result demonstrates that our method can fully exploit rich contexts of I3D features and generate better results. Compared with previous state-of-the-art methods, our method achieves the best performance. Meanwhile, the performance improvement for smaller AN is slightly more evident and our RTD does not employ any post-processing method such as NMS. As illustrated in Table \ref{ANet}, RTD-Net also achieves comparable results on ActivityNet-1.3. We analyze that annotations in ActivityNet-1.3 are relatively sparse and the average number of instance is 1.54 in a video (THUMOS14: 15.29). However, our RTD-Net models the pairwise context, and thus it generally requires multiple instances in each video.

\begin{figure}[t]
\begin{center}
\includegraphics[scale=0.33]{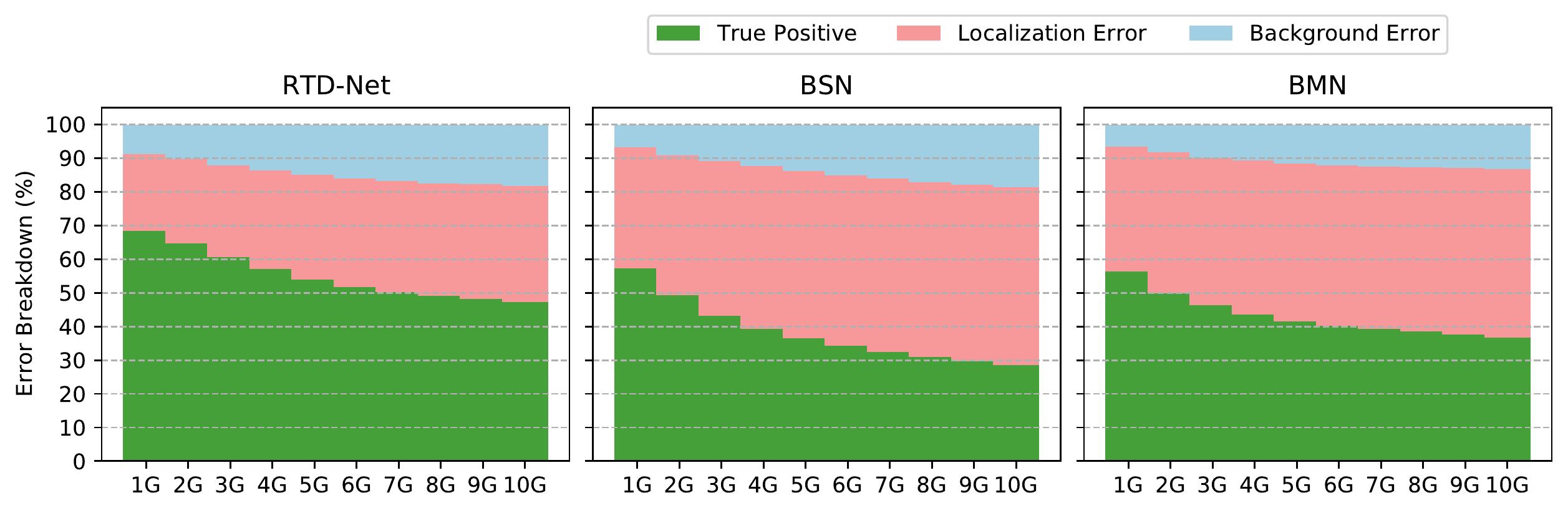}
\vspace{-1mm}
\caption{False positive profile of three proposal generation methods: RTD-Net, BSN and BMN. The three graphs demonstrate the FP error breakdown within the top 10-G (G = the number of ground truths) predictions per video. Maximum tIoU for localization error is set to 0.5.}
\label{fig:error_breakdown}
\end{center}
\vspace{-8mm}
\end{figure}

\noindent\textbf{In-depth analysis of RTD-Net proposals.}
We compare the results of RTD-Net with bottom-up methods BSN and BMN, via a false positive analysis. Inspired by~\cite{DBLP:conf/eccv/AlwasselHEG18}, we sort predictions by their scores and take the top-10G predictions per video. Two major errors in proposal generation task are discussed, localization error and background error.  Localization error is when a proposal is predicted as foreground, has a minimum tIoU of 0.1 but does not meet the tIoU threshold. Background error is when a proposal is predicted as foreground but its tIoU with ground truth instance is smaller than 0.1. In Figure~\ref{fig:error_breakdown}, we observe RTD-Net predictions has the most of true positive samples at every amount of predictions. The proportion of localization error in RTD-Net is notably smaller than those in BSN and BMN, confirming the overall precision of RTD predictions.

\begin{figure}[t]
\begin{center}
\includegraphics[scale=0.265]{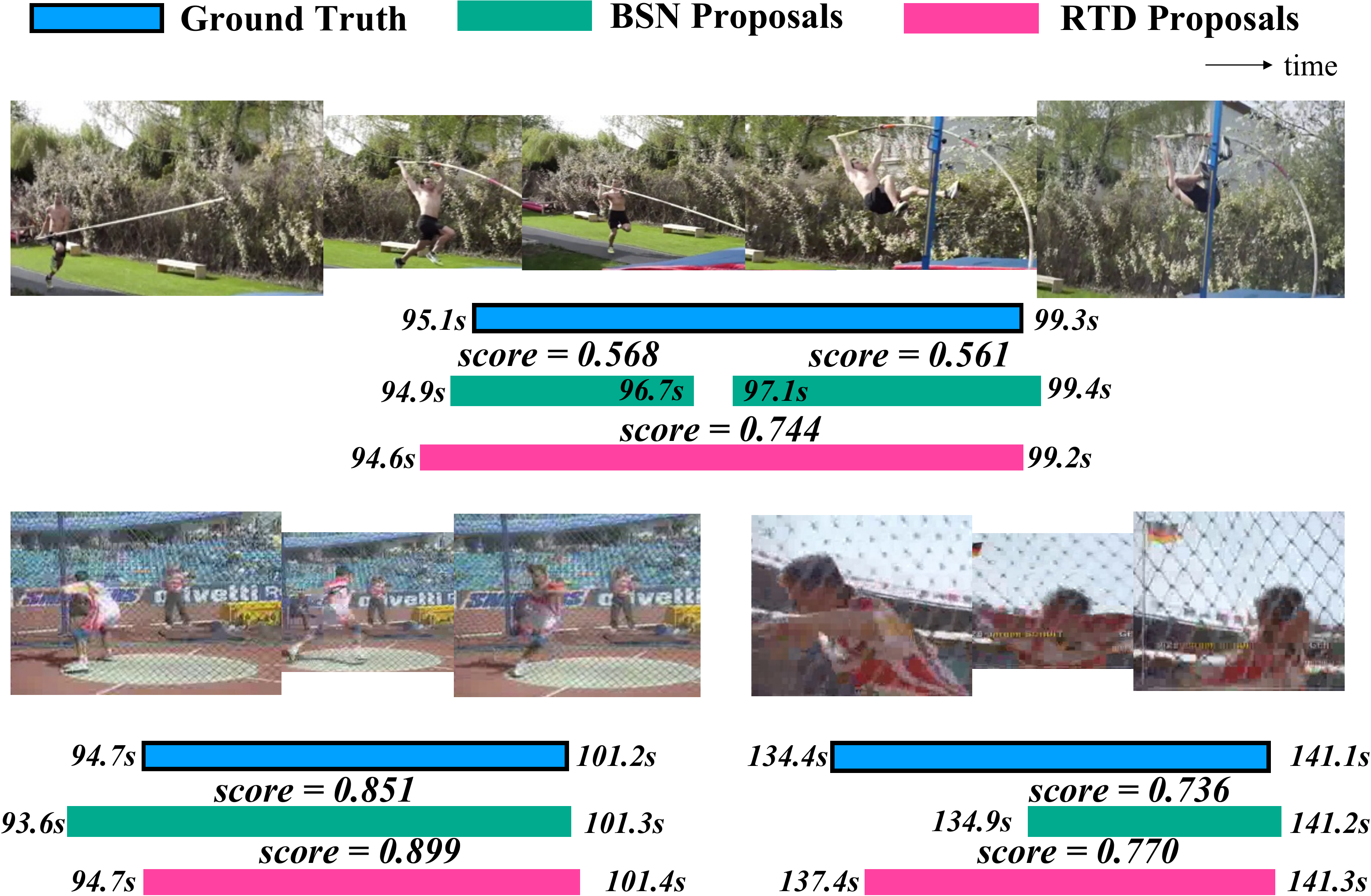}
\vspace{1mm}
\caption{Qualitative results of RTD-Net on THUMOS14. The proposals shown are the top predictions for corresponding groundtruths based on the scoring scheme for each model.}
\label{fig:thumos_qualitative}
\end{center}
\vspace{-8mm}
\end{figure}

We visualize qualitative results in Figure~\ref{fig:thumos_qualitative}. Specifically, while BSN makes two incomplete predictions for one action instance, RTD-Net predicts one complete proposal that accurately covers the whole action (the first row). Bottom-up methods like BSN only exploit context in a local window, hence they are unaware of similar features out of range. As a result, they are not robust to local noise and easily yield incomplete proposals. 
In the multi-instance setting (the second row), RTD-Net has better localization results with more precise boundaries or larger overlap with groundtruths. Benefit from global contextual information, RTD-Net is better aware of visual relationships between action proposals, and visual differences between foregrounds and backgrounds. Therefore, RTD-Net can easily distinguish between foreground and background segments, and localize proposals precisely.

\noindent\textbf{Time analysis in inference.} RTD-Net also has a notable advantage in inference speed. Compared with BSN, the inference time per sample of RTD-Net is much less (0.114s vs 5.804s, where 5.794s for BSN post-processing). Due to the direct proposal generation mechanism, RTD-Net is free of time-consuming post-processing methods such as non-maximum suppression. The experiment of inference speed is conducted on one RTX 2080Ti GPU. Detailed efficiency analysis is provided in Appendices C.

\begin{table}
\begin{center}
\caption{Ablation study on the boundary probability scaling factor on THUMOS14, measured by AR@AN.}
\vspace{1mm}
\scalebox{0.9}{
\begin{tabular}{c|cccc}
\toprule[1pt]
Scaling factor $\alpha$ & @50 & @100 & @200 & @500  \\ 
\hline
None & 36.22 &	45.38 &	52.62 &	59.61  \\
1 & 40.39 &	48.80 &	56.04 &	\textbf{63.41}  \\
2 & \textbf{41.52} & \textbf{49.32} & \textbf{56.41} & 62.91  \\
5 & 39.76 &	47.52 &	54.10 &	60.87  \\
\bottomrule[1pt]
\end{tabular}
}
\label{re-weight}
\end{center}
\vspace{-4mm}
\end{table}

\begin{table}
\begin{center}
\caption{Ablation study on feature encoders on THUMOS14, measured by AR@AN. }
\vspace{-2mm}
\scalebox{0.85}{
\begin{tabular}{cc|cccc}
\toprule[1pt]
Encoder & \tabincell{c}{Size of \\Receptive field*} & @50 & @100 & @200 & @500 \\ 
\hline
MLP & 64 & \textbf{41.52} & \textbf{49.32} & \textbf{56.41} & \textbf{62.91}   \\ 
Transformer & 64 & 33.69 & 40.36 & 46.33 & 52.38  \\ 
Transformer & 16 & 36.01 &	41.97 &	46.92 &	53.26  \\
\bottomrule[1pt]
\end{tabular}
}
\begin{tablenotes}
\footnotesize
\item * 'Size of Receptive field' means the temporal receptive field size of the input I3D features, the value is measured by frames per time step.
\end{tablenotes}
\label{feature}
\end{center}
\vspace{-6mm}
\end{table}

\subsection{Ablation Study}
\noindent\textbf{Study on scaling factor.} We re-weight video features with the predicted boundary probability sequence to enhance the features at possible boundary locations. Scaling factor $\alpha_r$ needs careful consideration, because it determines a probability threshold to decide the features at which location to enhance and to suppress, i.e. $\alpha_r = 2$ enhances features at locations with a boundary probability more than $0.5$ and suppresses those at locations with a probability less than $0.5$. Table \ref{re-weight} shows AR@AN on THUMOS14 dataset under different settings of the scaling factor. Comparing the results under different $\alpha_r$ settings, we observe that boundary-attentive representation boosts the performance up to \textbf{4\%} of average recall, and $\alpha_r
= 2$ maximizes the improvement.

\noindent\textbf{Study on feature encoders.} 
We analyze the design of the boundary-attentive module by experimenting on different encoder choices and input feature with different receptive field sizes. We compare results between MLP and Transformer encoder with the same high-level feature inputs. The first two rows in Table \ref{feature} show that MLP outperforms Transformer encoder by a \textbf{large margin} and we analyze that the performance drop might be caused by the over-smooth of self-attention in Transformer. To further investigate the performance, we experiment on the Transformer encoder with features of a smaller receptive field to reduce the over-smoothing effect, and the performance increases to around 36\%@50 but is still worse than our MLP encoder. 

\begin{table}[t]
\begin{center}
\caption{Ablation study on the relaxed matcher on THUMOS14 and ActivityNet-1.3, measured by AR@AN and AUC.}
\vspace{-2mm}
\scalebox{0.68}{
\begin{tabular}{c|cccc|ccc}
\toprule[1pt]
Relaxed matcher & @50 & @100 & @200 & @500 & AR@1 & AR@100 & AUC  \\ 
\hline
\xmark & 41.07 & 49.20 & 56.23 & 62.77 & 32.73 & 71.88 &  65.50  \\ 
\cmark & \textbf{41.52} & \textbf{49.32} & \textbf{56.41} & \textbf{62.91} & \textbf{33.05} & \textbf{73.21} & \textbf{65.78} \\
\bottomrule[1pt]
\end{tabular}
}
\label{relaxed_matcher}
\end{center}
\vspace{-6mm}
\end{table}

\begin{table}
\begin{center}
\caption{Ablation study on the tIoU guided ranking on THUMOS14, measured by AR@AN.}
\vspace{1mm}
\scalebox{0.85}{
\begin{tabular}{c|cccc}
\toprule[1pt]
Score & @50 & @100 & @200 & @500 \\ 
\hline
Classification& 41.08 & {49.03} & {56.07} &	\textbf{62.93}  \\ 
Classification~+~Completeness & 
\textbf{41.52} & \textbf{49.32} & \textbf{56.41} & {62.91}\\ 
\bottomrule[1pt]
\end{tabular}
}
\label{tIoU_guide}
\vspace{-2mm}
\end{center}
\end{table}

\noindent\textbf{Study on relaxed matcher.} With the relaxed matching criteria, some high-quality proposals, assigned negative in the original matcher, will become positive samples. As illustrated in Table \ref{relaxed_matcher}, the relaxed matcher can improve the AR and AUC metrics. In practice, we first train with the strict bipartite matching criteria to generate sparse predictions, then finetune with the relaxed matching scheme to improve the overall recall (more details in Appendices A.2).

\noindent\textbf{Study on completeness modeling.} The completeness head is designed to aid the binary classification score for a more reliable measure of predictions. We perform experiments on THUMOS14 testing set, and evaluate proposals in terms of AR@AN.  Table~\ref{tIoU_guide} reports the results of ablation study on the completeness head. We see that the combination classification and completeness scores  outperforms the result of simply using a classification score. We find that the estimated tIoU score is able to correct some well-predicted proposals but with a low classification score, and hence can boost the AR metrics especially at a smaller AN.

\begin{table}
\begin{center}
\caption{Temporal action detection results on the test set of THUMOS14 in terms of mAP at different tIoU thresholds. Proposals are combined with the classifiers of UntrimmedNet~\cite{DBLP:conf/cvpr/WangXLG17} and P-GCN~\cite{DBLP:conf/iccv/ZengHGTRZH19}.}

\scalebox{0.8}{
\begin{tabular}{ccccccc}
\toprule[1pt]
Method & Classifier & 0.7 & 0.6 & 0.5 & 0.4 & 0.3 \\ 
\hline
SST~\cite{DBLP:conf/bmvc/BuchEGFN17} & UNet & 4.7 &	10.9 & 20.0 & 31.5 & 41.2 \\
TURN~\cite{DBLP:conf/iccv/GaoYSCN17} & UNet & 6.3 & 14.1 &	24.5 & 35.3 & 46.3 \\
BSN~\cite{DBLP:conf/eccv/LinZSWY18} & UNet & 20.0 & 28.4 &	36.9 & 45.0 & 53.5 \\
MGG~\cite{DBLP:conf/cvpr/Liu0Z0C19}	& UNet & 21.3 & 29.5 &	37.4 & 46.8	& 53.9 \\
BMN~\cite{DBLP:conf/iccv/LinLLDW19}	& UNet & 20.5 & 29.7 &	38.8 & 47.4 & 56.0 \\
DBG~\cite{DBLP:conf/aaai/LinLWTLCWLHJ20} & UNet & 21.7 & 30.2 &	39.8 & 49.4 & 57.8 \\
G-TAD~\cite{DBLP:conf/cvpr/XuZRTG20} & UNet & 23.4 & 30.8 & 40.2 & 47.6 & 54.5 \\
BC-GNN~\cite{DBLP:conf/eccv/BaiWTYLL20} & UNet & 23.1 & 31.2 & 40.4 & 49.1 & 57.1 \\

\hline
RTD-Net & UNet & \textbf{25.0} &  \textbf{36.4} & \textbf{45.1} &	\textbf{53.1} & \textbf{58.5} \\
\hline
BSN~\cite{DBLP:conf/eccv/LinZSWY18} & P-GCN & - & - & 49.1 & 57.8 &	63.6 \\
G-TAD~\cite{DBLP:conf/cvpr/XuZRTG20} & P-GCN & 22.9 & 37.6 & 51.6 & 60.4 &	66.4 \\
\hline
RTD-Net & P-GCN & \textbf{23.7} &	\textbf{38.8}	& \textbf{51.9} & \textbf{62.3} & \textbf{68.3} \\
\bottomrule[1pt]
\end{tabular}
}
\label{TAD}
\end{center}
\vspace{-4mm}
\end{table}

\begin{table}
\begin{center}
\caption{Temporal action detection results on the validation set of ActivityNet-1.3 in terms of mAP at different tIoU thresholds. Proposals are combined with the classifier of UntrimmedNet~\cite{DBLP:conf/cvpr/WangXLG17}.}
\vspace{1mm}
\scalebox{0.9}{
\begin{tabular}{ccccccc}
\toprule[1pt]
Method & 0.95 & 0.75 & 0.5 & Average \\ 
\hline
SCC~\cite{DBLP:conf/cvpr/HeilbronBEG17} & 4.70 &	17.90 & 40.00 & 21.70\\
CDC~\cite{DBLP:conf/cvpr/ShouCZMC17} & 0.21 & 25.88 & 43.83 & 22.77 \\
SSN~\cite{DBLP:conf/iccv/ZhaoXWWTL17}	& 5.49 & 23.48 & 39.12 & 23.98 \\
Lin et al.\cite{DBLP:journals/corr/LinZS17}	& 7.09 & 29.65 & 44.39 & 29.17\\
BSN~\cite{DBLP:conf/eccv/LinZSWY18}	& 8.02 & 29.96 & 46.45 & 30.03\\
BMN~\cite{DBLP:conf/iccv/LinLLDW19}	& 8.29 & \textbf{34.78} & \textbf{50.07} & \textbf{33.85}\\
\hline
RTD-Net & \textbf{8.61} & 30.68 & 47.21 & 30.83 \\
\bottomrule[1pt]
\end{tabular}
}
\label{TAD_ANet}
\end{center}
\vspace{-8mm}
\end{table}

\subsection{Action Detection with RTD proposals}
\noindent\textbf{Evaluation metrics.} To evaluate the results of temporal action detection task, we calculate Mean Average Precision (mAP). On THUMOS14, mAP with tIoU thresholds set [0.3 : 0.1 : 0.7] are calculated. On ActivityNet-1.3, mAP with tIoU thresholds set \{0.5, 0.75, 0.95\} and average mAP with tIoU thresholds set [0.5 : 0.05 : 0.95] are reported.

\noindent\textbf{Comparison with state-of-the-art methods.} To evaluate the quality of our proposals for action detection, we follow a two-stage temporal action detection pipeline. First, we generate a set of action proposals for each video with our RTD-Net and keep top-200 and top-100 proposals for subsequent detection on THUMOS14 and ActivityNet-1.3. Then we score each proposal with two specific strategies. One strategy is using a global classification score from UntrimmedNet~\cite{DBLP:conf/cvpr/WangXLG17} and keeping top-2 predicted labels for each video. Then, we assign the classification score to each proposal and use fusion proposal confidence score and global classification score as detection score. The other strategy is that we employ the proposal-level classifier P-GCN~\cite{DBLP:conf/iccv/ZengHGTRZH19} to predict action labels for each proposal and use the predicted score for evaluation.

The result on THUMOS14 is shown in Table \ref{TAD} and our RTD-Net based detection outperforms other state-of-the-art methods especially under \textbf{high tIoU} settings, which indicates that proposals generated by RTD-Net are more accurate. When combined with P-GCN classifier, our method achieves mAP improvements over other proposal generation methods such as BSN~\cite{DBLP:conf/eccv/LinZSWY18} and G-TAD~\cite{DBLP:conf/cvpr/XuZRTG20} at all tIoU thresholds. This experiment demonstrates that \emph{RTD proposals are able to boost the performance of temporal action detection task}. As Table~\ref{TAD_ANet} illustrates, we achieve comparable results on ActivityNet-1.3. BSN and BMN~\cite{DBLP:conf/iccv/LinLLDW19} predict a large number of proposals (nearly 900 proposals per video) and select top-100 of them, while RTD-Net only makes 100 predictions. Compared with BSN and BMN, RTD-Net improves mAP under \textbf{high tIoU} settings (tIoU = 0.95), since RTD-Net proposals have more precise boundaries.

\section{Conclusion}
In this paper, we have proposed a simple pipeline for direct action proposal generation by re-purposing a Transformer-alike architecture. To bridge the essential difference between videos and images, we introduce three important improvements over the original DETR framework, namely a boundary attentive representation, a relaxed Transformer decoder (RTD), and a three-branch prediction head design. Thanks to the parallel decoding of multiple proposals with explicit context modeling, our RTD-Net outperforms the previous state-of-the-art methods in temporal action proposal generation task on THUMOS14 and also yields a superior performance for action detection on this dataset. In addition, free of NMS post-processing, our detection pipeline is more efficient than previous methods. 

\vspace{-2mm}
\paragraph {\bf Acknowledgements.} \small This work is supported by National Natural Science Foundation of China (No. 62076119, No. 61921006), Program for Innovative Talents and Entrepreneur in Jiangsu Province, and Collaborative Innovation Center of Novel Software Technology and Industrialization.

\newpage
\appendix

\setcounter{equation}{0}
\setcounter{section}{0}
\setcounter{subsection}{0}
\setcounter{table}{0}
\setcounter{figure}{0}

\renewcommand\thetable{\Alph{table}}
\normalsize
\renewcommand\thesection{\Alph{section}}
\renewcommand\thefigure{\Alph{figure}}

\section{Additional Ablation Studies}
\subsection{Boundary Attentive Module}
\noindent\textbf{Study on attentive representations.} 
To further analyze the design of the boundary-attentive module, we perform ablations on projection placement (i.e., location of MLP) and boundary enhancement methods. For projection placement, we consider the alternatives of pre-enhancement and post-enhancement. For boundary enhancement methods, we also tried concatenating boundary weights with features along the channel dimension. Results of Table~\ref{proj} show that pre-enhancement projection has a better performance and multiplication enhancement outperforms concatenation enhancement. 
We analyze that pre-enhancement projection provides a more compact representation for boundary enhancement and attention introduces a more direct and explicit feature enhancement strategy. 

\begin{table}[!h]
\begin{center}
\caption{Ablation study on MLP encoders on THUMOS14, measured by AR@AN.}
\vspace{-3.5mm}
\scalebox{0.9}{
\begin{tabular}{cc|cccc}
\toprule[1pt]
\tabincell{c}{Projection \\ placement} & \tabincell{c}{Boundary \\ enhancement} & @50 & @100 & @200 & @500 \\ 
\hline
Pre & multiply & \textbf{41.52} & \textbf{49.32} & \textbf{56.41} & \textbf{62.91} \\
Post & multiply & 38.81 & 47.36 & 54.86 & 62.30  \\ 
Pre & concat & 37.92 & 45.33 & 52.12 & 60.86  \\
\bottomrule[1pt]
\end{tabular}
}
\label{proj}
\end{center}
\vspace{-2mm}
\end{table}

\noindent\textbf{Study on temporal positional embedding in encoder.} 
In this section, we show the importance of temporal positional embedding in the boundary attentive module. We experiment with removing positional embedding at MLP encoder or directly adding it into encoder. We contend that concatenating positional embedding with video features explicitly gives the encoded features the relative order of the sequence, and simplifies the difficulty of proposal generation by having temporal locations encoded in the features. The results in Table \ref{position_embed} show that the model performance decreases by 4.4\% on AR@50, without temporal positional embedding in encoder.

\begin{table}[!h]
\begin{center}
\caption{Ablation study on position embedding of MLP encoder on THUMOS14, measured by AR@AN.}
\scalebox{0.8}{
\begin{tabular}{c|ccccc}
\toprule[1pt]
\tabincell{c}{Positional embedding \\ in encoder } & @50 & @100 & @200 & @500 \\ 
\hline
w/o  & 37.07 & 45.05 & 51.58 & 58.31 \\ 
w/ & \textbf{41.52} & \textbf{49.32} & \textbf{56.41} & \textbf{62.91} \\
\bottomrule[1pt]
\end{tabular}
}
\label{position_embed}
\end{center}
\vspace{-2mm}
\end{table}

\noindent\textbf{Effect of feature receptive field on MLP encoder.}
Table~\ref{MLP_feature} is an extension of Table 4 in Section 4.3 to prove that the over-smoothing effect of encoder self-attention causes performance drop. We extend our experiment to alleviate the possibility that features with smaller receptive field boosts the performance in general. By comparing MLP encoder performance of input features with receptive field size of 16 and 64, we conclude that smaller receptive field would decrease the performance of MLP encoder. The increase of performance with Transformer encoder is because that smaller receptive field reduces the over-smoothing effect for Transformer encoder.
\begin{table}[!h]
\begin{center}
\caption{Ablation study on the effect of feature receptive field on MLP encoder on THUMOS14, measured by AR@AN.}
\scalebox{0.9}{
\begin{tabular}{c|cccc}
\toprule[1pt]
\tabincell{c}{Size of \\ Receptive field } & @50 & @100 & @200 & @500 \\ 
\hline
64 & \textbf{41.52} & \textbf{49.32} & \textbf{56.41} & \textbf{62.91} \\
16  & 39.56 & 47.36 & 53.82 & 60.47  \\ 
\bottomrule[1pt]
\end{tabular}
}
\label{MLP_feature}
\end{center}
\vspace{-2mm}
\end{table}


\subsection{Relaxed Transformer Decoder}
\noindent\textbf{Study on the relaxation mechanism.}
We present the two-step \textbf{“top-1 to top-k”}  matching scheme. In our strategy, we first train with the strict bipartite matching criteria to generate sparse predictions, then fine-tune with the relaxed matching scheme to improve the overall recall. The first step of our strategy is {\em necessary} because it makes the positive samples sparsely distributed and minor-overlapped, thus the model is free of NMS.

In the fine-tuning phase, we freeze the modules except for binary classification and boundary embeddings. Specifically, we calculate tIoU between targets and predictions, and employ three different settings of the relaxation mechanism. \emph{First}, we mark predictions with tIoU higher than a threshold as positive samples and get an updated matching permutation $\sigma’$. We calculate both classification and localization loss according to the updated assignment $\sigma’$. \emph{Second}, only loss for the binary classification head is calculated with $\sigma’$. The target of this relaxation setting is to improve the quality measurement (confidence) of positive (but not optimal) proposals, and stabilize the distribution of optimal predictions. The \emph{last} one is assigning the closest prediction of each groundtruth as positive elements (predictions of bipartite matching are not necessarily the geometrically closest), and calculate losses on this updated assignment $\sigma{’}{’}$. As Table~\ref{relax} illustrates, the results of all three settings are close, demonstrating the influence of the relaxation mechanism is robust to settings (rule and scope).

With the relaxation mechanism, our model witnesses an evident improvement on AR and AUC. With the optimal bipartite matching, RTD-Net predicts proposals of bipartite matching (top-1 proposals) well, while it suppresses several other predictions around the groundtruth (top-k proposals), which results in a decrease of AR at large AN and overall AUC. In the fine-tuning phase, our model improves the scoring of top-k proposals with the relaxation mechanism, and the performance of top-1 proposals is not affected. As a result, the relaxation mechanism boosts the overall performance of RTD-Net.

Similar to us,~\cite{DBLP:journals/corr/abs-2101-11782} exploits a “stop-grad” operation, namely they freeze the FCOS detector~\cite{DBLP:conf/iccv/TianSCH19} and train their PSS head in the fine-tuning phase. The difference is that~\cite{DBLP:journals/corr/abs-2101-11782} firstly makes top-k predictions well and then learns to predict top-1 proposals. RTD-Net exploits a \textbf{“top-1 to top-k”} strategy, while~\cite{DBLP:journals/corr/abs-2101-11782} leverages a \textbf{“top-k to top-1”} scheme. Both of them aim to optimize the procedure of label assignment at the cost of removing heuristic NMS, and markedly reduce the inference time.  

\begin{table}[!h]
\begin{center}
\caption{Ablation study on the rule of relaxation mechanism on ActivityNet-1.3 validation set, measured by AR@AN and AUC.}
\scalebox{0.8}{
\begin{tabular}{c|c|ccc}
\toprule[1pt]
Rule & Scope & AR@1 & AR@100 & AUC \\ 
\hline
None & None & 32.73 & 71.88 & 65.50 \\ 
threshold & cls + loc & 33.05 & 73.21 & \textbf{65.78} \\
threshold & cls & \textbf{33.10} & 73.12 & 65.77 \\
top1 & cls + loc & 32.95 & \textbf{73.25} & 65.77 \\ 
\bottomrule[1pt]
\end{tabular}
}
\label{relax}
\end{center}
\vspace{-2mm}
\end{table}

\noindent\textbf{Study on temporal positional embedding in decoder.}
Explicit temporal positional embedding also plays a key role in the relaxed transformer decoder. We experiment with no positional embedding, add positional embedding at encoder-decoder attention input and similar to detr, add positional embedding only at attention. As shown in Table~\ref{decoder_position_embed}, adding positional embedding at attention achieves the best performance. RTD-Net achieves 37.43\% on AR@50 without positional embedding in the decoder, which decreases by about 4\%. Adding positional embedding at input causes performance drop as well, by 2.0\% on AR@50. 

\begin{table}[!h]
\begin{center}
\caption{Ablation study on position embedding of transformer decoder on THUMOS14, measured by AR@AN.}
\scalebox{0.8}{
\begin{tabular}{c|cccc}
\toprule[1pt]
\tabincell{c}{Positional embedding} & @50 & @100 & @200 & @500 \\ 
\hline
None  & 37.43 & 46.01 & 53.90 & 61.32 \\ 
At input  & 39.53 & 47.13 & 53.83 & 61.67 \\ 
At attn. & \textbf{41.52} & \textbf{49.32} & \textbf{56.41} & \textbf{62.91} \\
\bottomrule[1pt]
\end{tabular}
}
\label{decoder_position_embed}
\end{center}
\vspace{-2mm}
\end{table}

\noindent\textbf{Study on the number of decoder layers.} We conduct experiments on the number of decoder layers and the results are displayed in Table \ref{decoder_layers}. RTD-Net achieves the best performance with 6 decoder layers, in terms of AR@AN. When the number of decoder layers increases from 1 to 2, it improves AR@50 by around 6.2, but this improvement decreases to 1.8 when the number of decoder layers increases from 2 to 3.

\begin{table}[!h]
\begin{center}
\caption{Ablation study on the number of decoder layers on THUMOS14, measured by AR@AN.}
\scalebox{0.8}{
\begin{tabular}{c|cccc}
\toprule[1pt]
\tabincell{c}{Number of decoder layers} & @50 & @100 & @200 & @500 \\ 
\hline
1 & 32.76 & 42.93 & 51.09 & 58.19 \\ 
2 & 38.92 & 47.47 & 53.14 & 60.11 \\ 
3 & 40.71 & 47.57 & 53.84 & 60.30 \\ 
6 & \textbf{41.52} & \textbf{49.32} & \textbf{56.41} & \textbf{62.91} \\
9 & 38.36 & 46.70 & 53.70 & 60.01 \\ 
\bottomrule[1pt]
\end{tabular}
}
\label{decoder_layers}
\end{center}
\vspace{-2mm}
\end{table}

\subsection{Non-Maximum Suppression}
In Table~\ref{nms}, we conduct experiments on RTD-Net with and without NMS, and observe similar results. NMS is not necessary in RTD-Net because the predictions are relatively sparse and minor-overlapped with our two-step training strategy (details in Appendices A.2). In contrast, BSN~\cite{DBLP:conf/eccv/LinZSWY18} and BMN~\cite{DBLP:conf/iccv/LinLLDW19} generate highly overlapped proposals with similar confidence, as shown in Figure \ref{fig:anet_qualitative} of Appendices. Therefore, NMS is needed for these dense proposal generators to suppress such proposals.

\begin{table}[h]
\begin{center}
 \vspace{-1mm}
\caption{Ablation study on non-maximum suppression on THUMOS14, measured by AR@AN.}
\vspace{1mm}
\scalebox{0.8}{
\begin{tabular}{c|cccc}
\toprule[1pt]
Method & @50 & @100 & @200 & @500 \\ 
\hline
RTD-Net & 41.52 & 49.32 & \textbf{56.41} & \textbf{62.91}  \\ 
RTD-Net+SNMS & \textbf{42.02} & \textbf{49.40} & 54.98 & 61.16\\ 
\bottomrule[1pt]
\end{tabular}
}
\label{nms}
\vspace{-4mm}
\end{center}
\end{table}

\section{Visualization}

\begin{figure*}
\begin{center}
\includegraphics[scale=0.5]{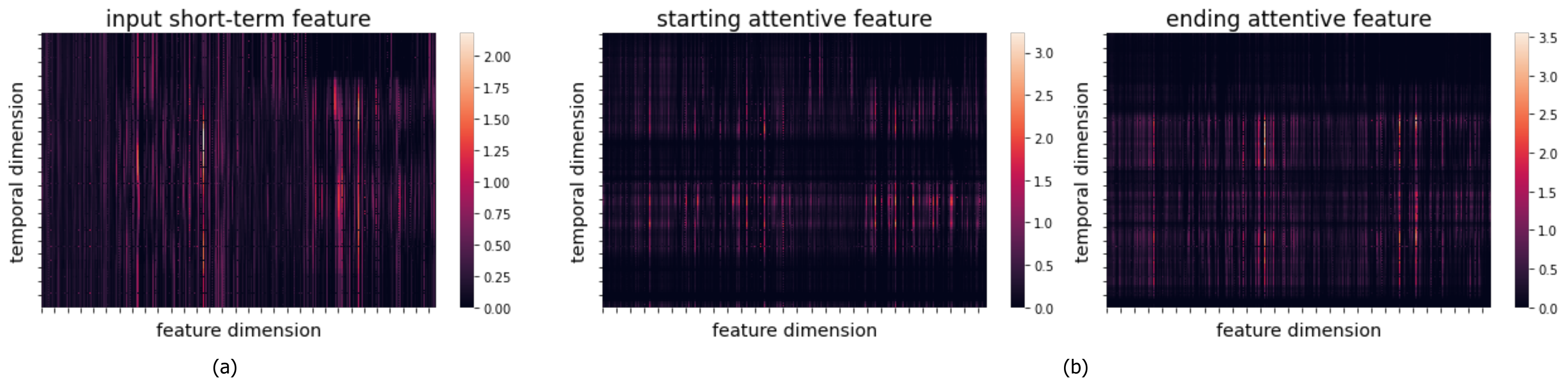}
\end{center}
\vspace{-3mm}
   \caption{(a) is visualization of input short-term feature of a randomly sampled video segment, this feature has a receptive field of 64 frames; (b) is visualization of starting and ending attentive features. Best viewed in color.}
\label{fig:input_fts}
\label{Input_Features}
\vspace{-2mm}
\end{figure*}

\begin{figure}
\begin{center}
\includegraphics[scale=0.4]{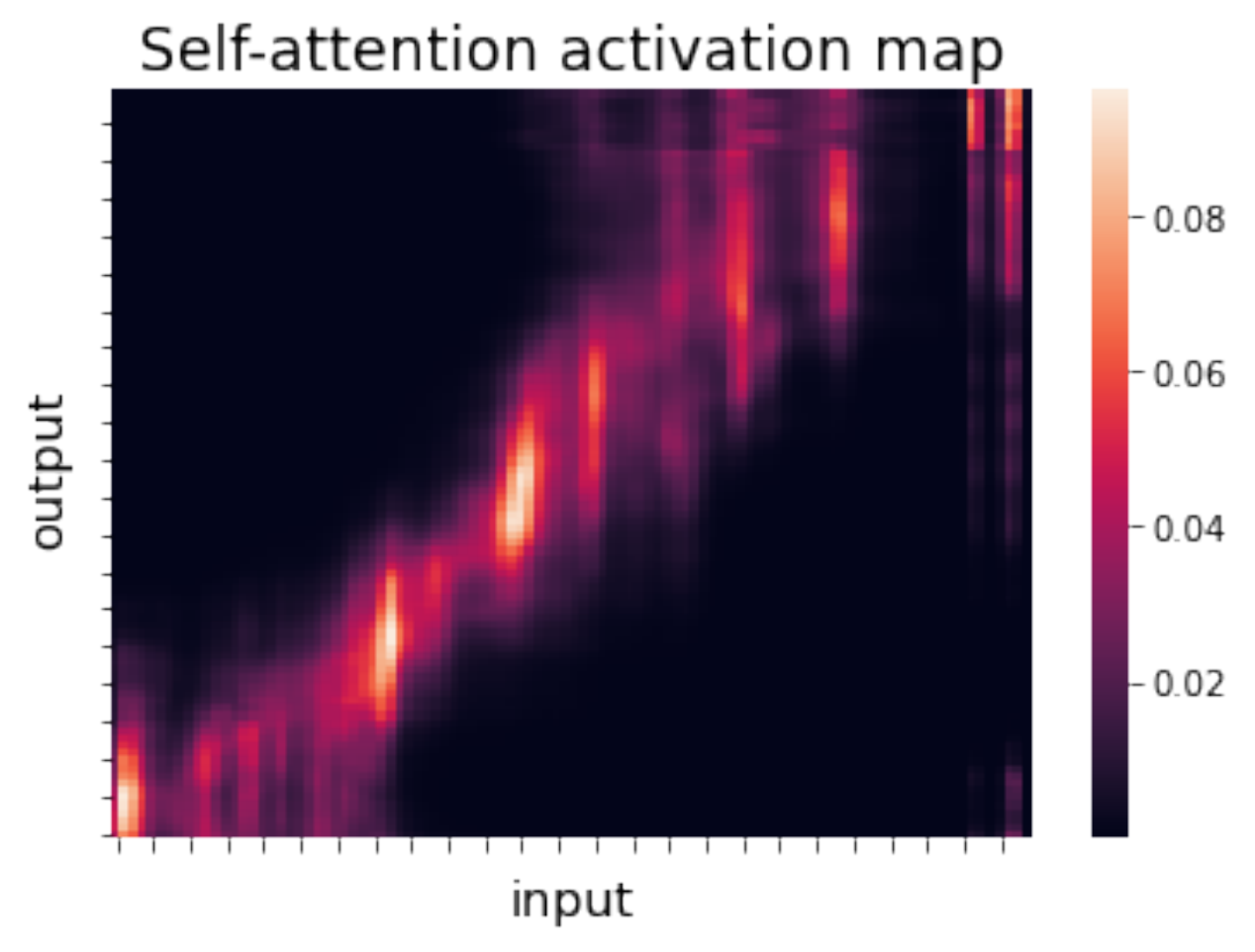}
\end{center}
\vspace{-2mm}
   \caption{ Visualization of self-attention activation map in Transformer encoder. Best viewed in color.}
\label{fig:selfattn}
\label{Self_Attn}
\vspace{-2mm}
\end{figure}

\noindent\textbf{Visualization of boundary-attentive representations.}
Figure~\ref{Input_Features}(a) shows the pattern for input video feature. Vertical line patterns are visible in input features, indicating different temporal locations sharing similar feature representation. That is the slowness phenomena that we discover in short-term video features. To alleviate this slowness, we explicitly multiply starting and ending attentive scores with features. Figure~\ref{Input_Features}(b) illustrates the starting and ending attentive feature. We observe the aforementioned vertical line patterns are broken by horizontal darker line patterns, indicating that effectiveness of boundary information in representation enhancement.

\noindent\textbf{Analysis on the over-smoothing effect.}
We further explore the reason for the over-smoothing effect with self-attention mechanism of the transformer encoder. Figure~\ref{Self_Attn} shows the self-attention map of a sample from THUMOS14~\cite{THUMOS14}. The x-axis is the input temporal locations, and the y-axis is the output temporal locations. A diagonal activation pattern is observed in Figure~\ref{Self_Attn}, with many short vertical line patterns visible around the diagonal activation. The vertical patterns indicate that many different output locations share the same input activation, which result in the over-smoothing effect. The input short-term feature already has the problem of slowness, adding temporal attention to this feature would aggravate the slowness and result in weaker performance.

\begin{figure*}
\begin{center}
\includegraphics[scale=0.45]{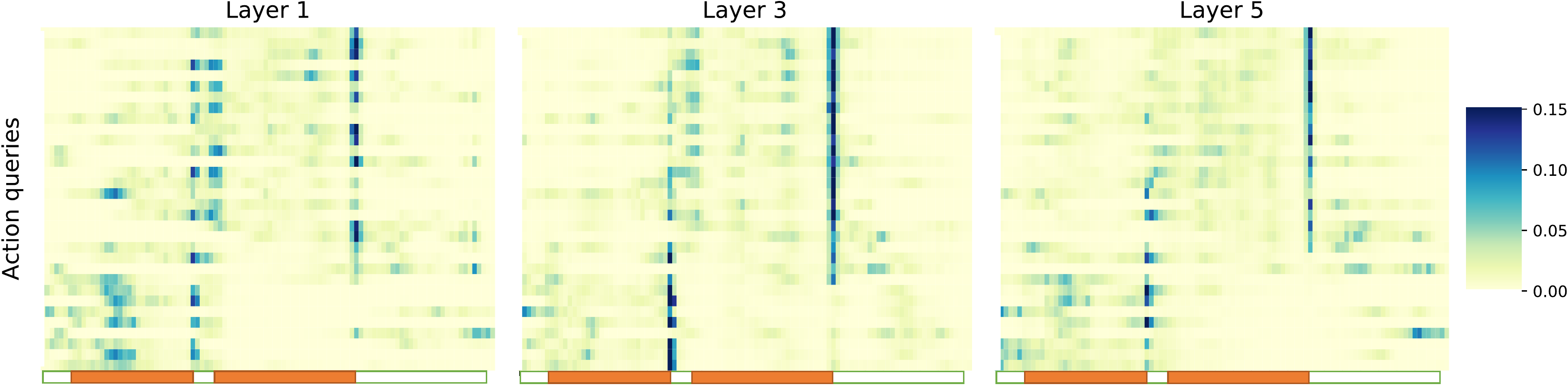}
\end{center}
\vspace{-2mm}
   \caption{ Visualization of encoder-decoder attention activation map, averaged among multiple heads. The y-axis is action queries and the x-axis represents time steps from encoder features. From yellow to blue represents the intensity of activation, the bluer the stronger the activation. The white and orange bar underneath the x-axis demonstrates groundtruth instances in this snippet. The orange part represents action and the rest represents background. Best viewed in color.}
\label{fig:decoder_crossattn}
\vspace{-2mm}
\end{figure*}

\noindent\textbf{Visualization of decoder attention maps.}
In this subsection, we present the activation map from self-attention layer and encoder-decoder attention layer in RTD decoder layers. Figure~\ref{fig:decoder_crossattn} shows the $N_Q \times N_T$ ($N_T$ is the number of time steps in each snippet, $N_Q$ is the number of queries predicted for each snippet) encoder-decoder activation map from Layer 1, 3 and 5 (last) of decoder layers from a randomly selected video snippet. Vertical patterns are visible in these activation maps. Each blue vertical beam corresponds to the ending of an action instance, which indicates that proposal queries are more focused on the features from the ending region of an action. 

\begin{figure}
\begin{center}
\includegraphics[scale=0.45]{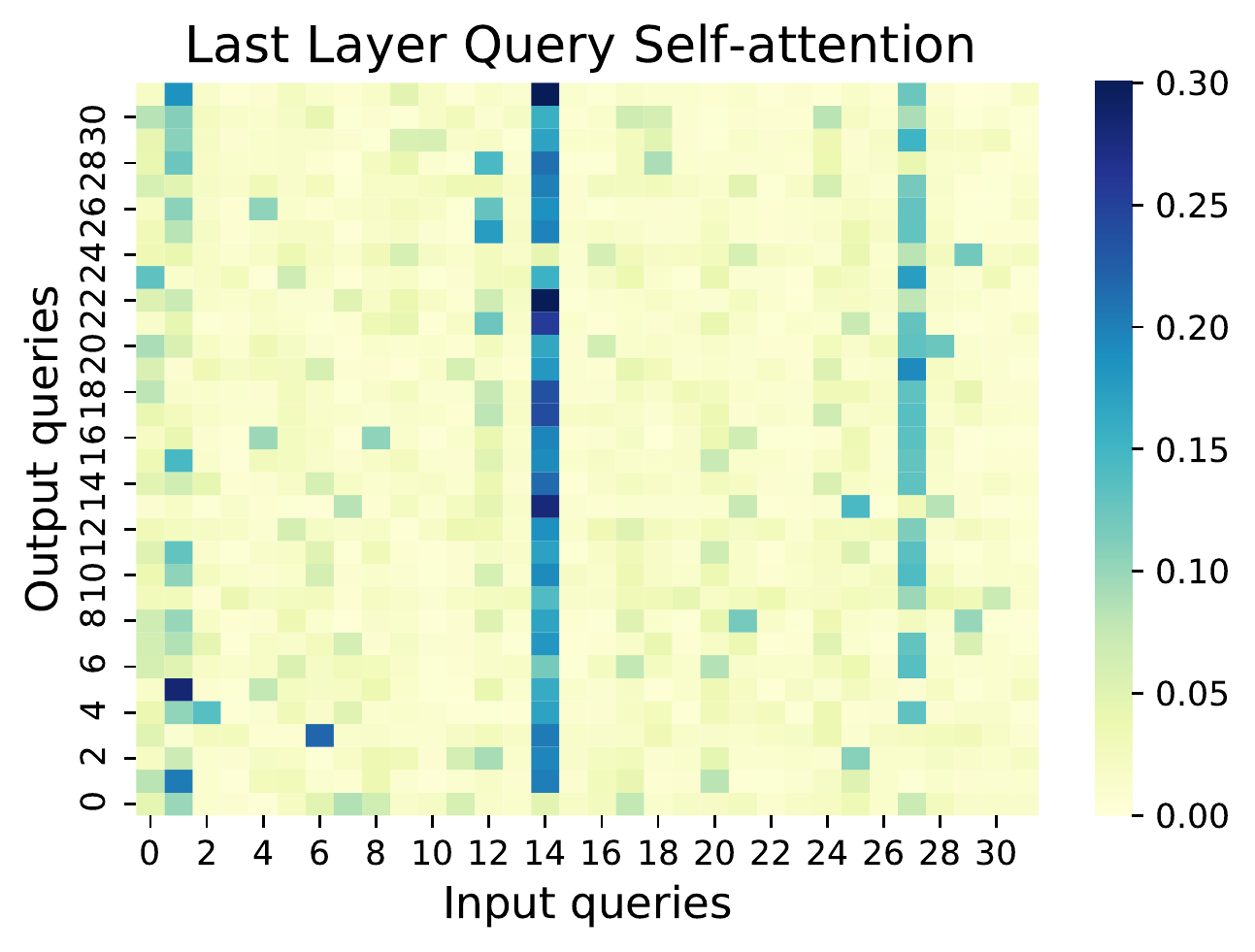}
\end{center}
\vspace{-2mm}
   \caption{Visualization of the self-attention layer in the last layer of Transformer decoder, averaged among multiple heads. Best viewed in color.}
\label{fig:query_selfattn}
\vspace{-2mm}
\end{figure}

Figure~\ref{fig:query_selfattn} shows the $N_Q \times N_Q$ query self-attention activation map from the last layer of decoder. High activations are visible along the y-axis, indicating that proposal queries are keen at learning from some well predicted queries (eg. 1st, 14th and 27th) at inter-proposal modeling. The 14th query in Figure~\ref{fig:query_selfattn} is the highest ranked and also a well-predicted proposal in results.

\section{Additional Comparisons with SOTA }
\begin{figure}
\begin{center}
\includegraphics[scale=0.45]{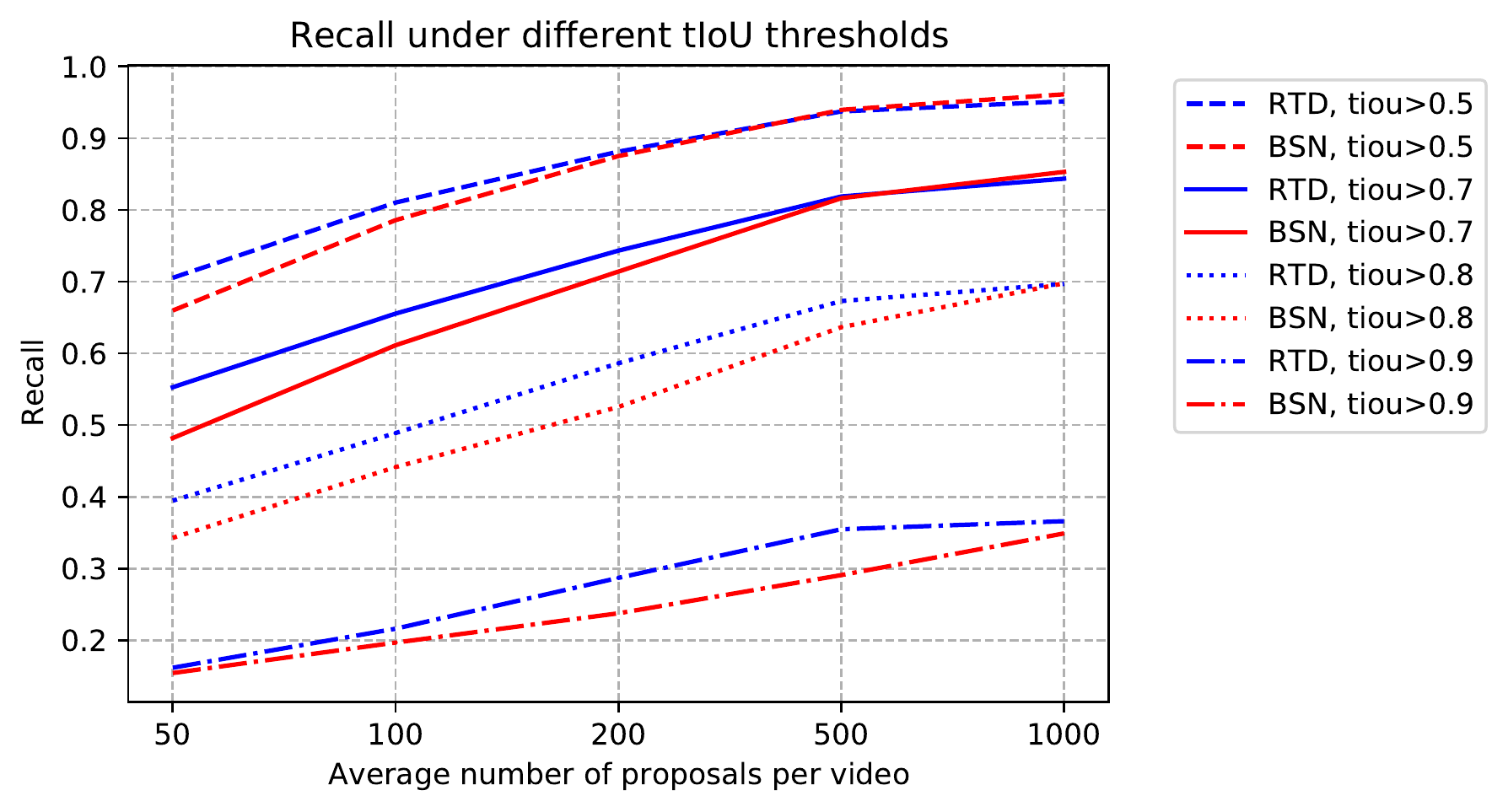}
\end{center}
\vspace{-2mm}
   \caption{Visualization of Average Recall at different proposal numbers under all tIoU thresholds. Best viewed in color.}
\label{fig:ar_curve}
\vspace{-2mm}
\end{figure}

\noindent\textbf{AR curves under all tIoU thresholds.}
RTD-Net generates more precise and more complete proposals, compared with previous methods. We compare RTD-Net with bottom-up method BSN under different tIoU thresholds for recall. In Figure~\ref{fig:ar_curve}, we demonstrate that: 1) RTD-Net outperforms BSN under every tIoU threshold, especially at smaller number of proposal conditions. 2) RTD-Net outperforms BSN under high tIoU thresholds, indicating that when the true positive standard is strict with localization, RTD-Net still achieves higher recall with better localized predictions.

\begin{figure}
\begin{center}
\includegraphics[scale=0.28]{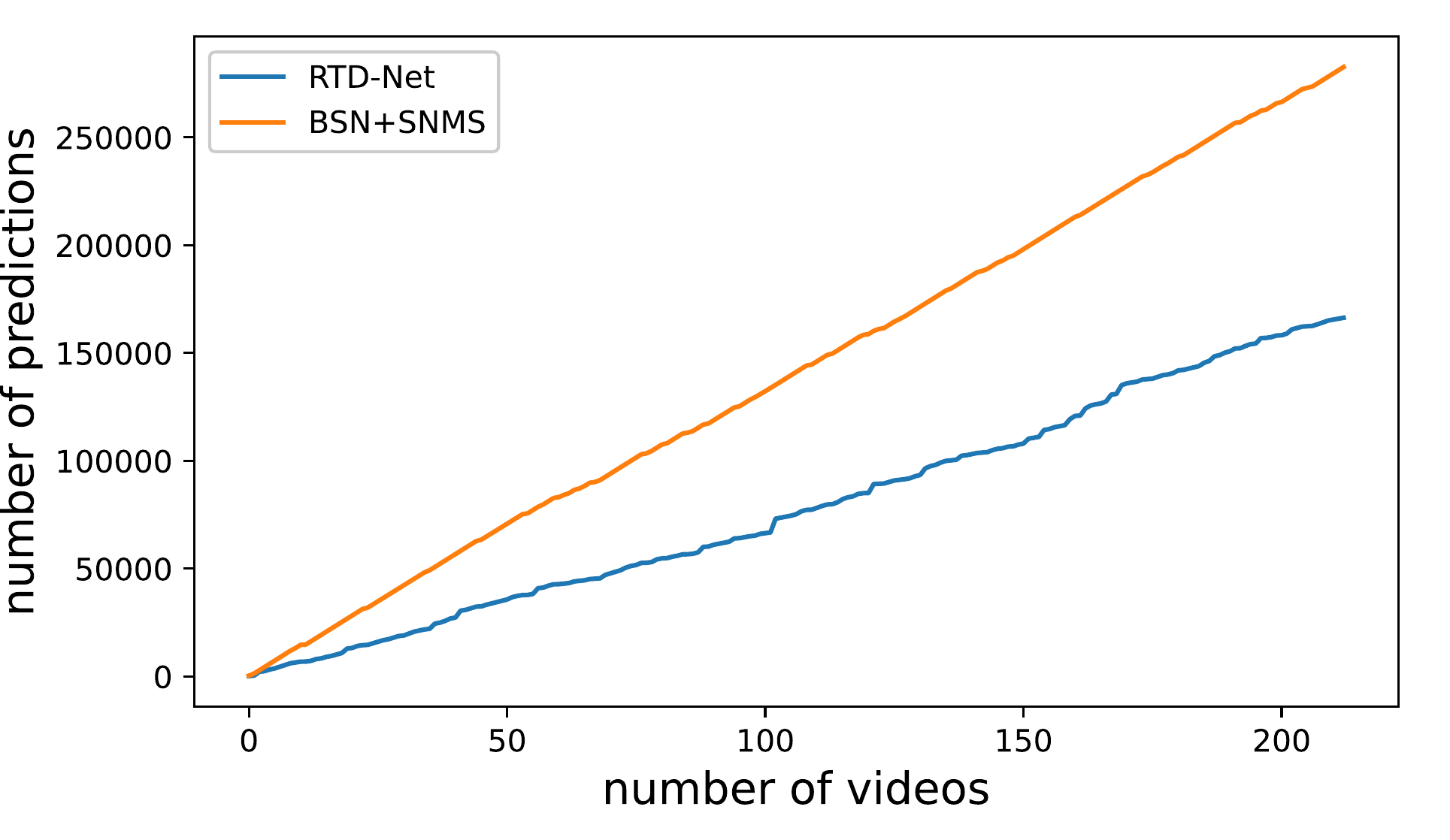}
\end{center}
\vspace{-3mm}
   \caption{Comparison of number of proposals between RTD-Net and BSN.}
   \vspace{-3mm}
\label{fig:num_proposals}
\end{figure}

\noindent\textbf{Efficiency Analysis.} Our RTD-Net only presents the transformer decoder, while keeping the original MLP encoder for feature extraction. Therefore, our encoder is with linear run-time and memory complexity. Our decoder uses cross attention and the complexity is $O(N_T \times N_Q)$. In practice, $N_Q$ could be smaller than sequence length. In our experiment, we found our method uses 1,519 MB GPU memory while existing SOTA methods such as BMN uses 7,152 MB. In addition, we provide a run-time breakdown for RTD-Net and BSN in Table \ref{runtime_breakdown}. We infer with 3-minute video input on one RTX 2080-Ti GPU. We follow \cite{DBLP:conf/eccv/LinZSWY18, DBLP:conf/iccv/LinLLDW19} to exclude the backbone feature extractor. It is noted that, for a 3-minute video, RTD-Net predicts 640 proposals without any post-processing module while BSN outputs about 3k predictions for the time-consuming SNMS post-processing.

\begin{table}
\caption{Run-time breakdown analysis of RTD-Net and BSN.}
\centering

\subtable[RTD-Net]{
\scalebox{0.7}{
\begin{tabular}{c|cccc}
\hline
 & Boundary-probability & MLP & Transformer & Three-branch\\ 
  & predictor + re-weight & encoder & decoder & Head \\ 
\hline
RTD-Net & 49.29ms & 0.32ms & 8.97ms & 0.89ms \\ 
 \hline
\end{tabular}
}
}
\qquad

\subtable[BSN]{       
\scalebox{0.7}{
\begin{tabular}{c|cccc}
\hline
 & TEM & PGM & PEM & SNMS\\ 
\hline
BSN & 53.23ms & 243.79ms & 7.68ms & 6026.34ms \\ 
 \hline
\end{tabular}
}
}
\label{runtime_breakdown}
\vspace{-2mm}
\end{table}

RTD-Net directly generates high-quality proposals with a smaller number of predictions. Due to the pair-wise modeling in our decoder, our predictions do not suffer from the flooding of redundant, highly-overlapping proposals. As shown in Figure~\ref{fig:num_proposals}, RTD-Net predicts fewer proposals than BSN~\cite{DBLP:conf/eccv/LinZSWY18}, but still achieves higher average recall under all metrics on THUMOS14.

\noindent\textbf{Generalizability of proposals.} The ability of generating high quality proposals for unseen action categories is an important property of a  temporal action proposal generation method. Following BSN~\cite{DBLP:conf/eccv/LinZSWY18} and BMN~\cite{ DBLP:conf/iccv/LinLLDW19}, we choose two non-overlapped action subsets: “Sports, Exercise, and Recreation” and “Socializing, Relaxing, and Leisure” of ActivityNet-1.3, as \emph{seen} and \emph{unseen} subsets separately. \emph{Seen} subset contains 87 action classes with 4455 training and 2198 validation videos, and \emph{unseen} subset contains 38 action classes with 1903 training and 896 validation videos. Based on I3D features, we train RTD-Net with \emph{seen} and \emph{seen+unseen} training videos separately, and evaluate on both \emph{seen} and \emph{unseen} validation videos. Results in Table~\ref{generalizability} demonstrate that the performance remains competitive in unseen categories, suggesting that RTD-Net achieves great generalizability to generate high quality proposals for unseen classes, and is able to predict accurate temporal action proposals regardless of semantics.

\begin{table}[h]
\begin{center}
\vspace{-2mm}
\caption{\small Generalization evaluation of RTD-Net on ActivityNet-1.3.}
\vspace{1mm}
\scalebox{0.7}{
\begin{tabular}{c|cccc}
\toprule[1pt]
& \multicolumn{2}{c} {\emph{Seen}(val)} & \multicolumn{2}{c} {\emph{Unseen}(val)}\\
\hline
 & AR@100 & AUC & AR@100 & AUC \\ 
\hline
\emph{Seen+Unseen}(train) & 70.25 & 62.66 & 73.09 & 65.52  \\ 
\emph{Seen}(train) & 69.80 & 61.32 & \textbf{72.27} & \textbf{64.54}\\ 
\bottomrule[1pt]
\end{tabular}
}
\label{generalizability}
\vspace{-2mm}
\end{center}
\end{table}

\begin{figure}[t]
\begin{center}
\includegraphics[scale=0.17]{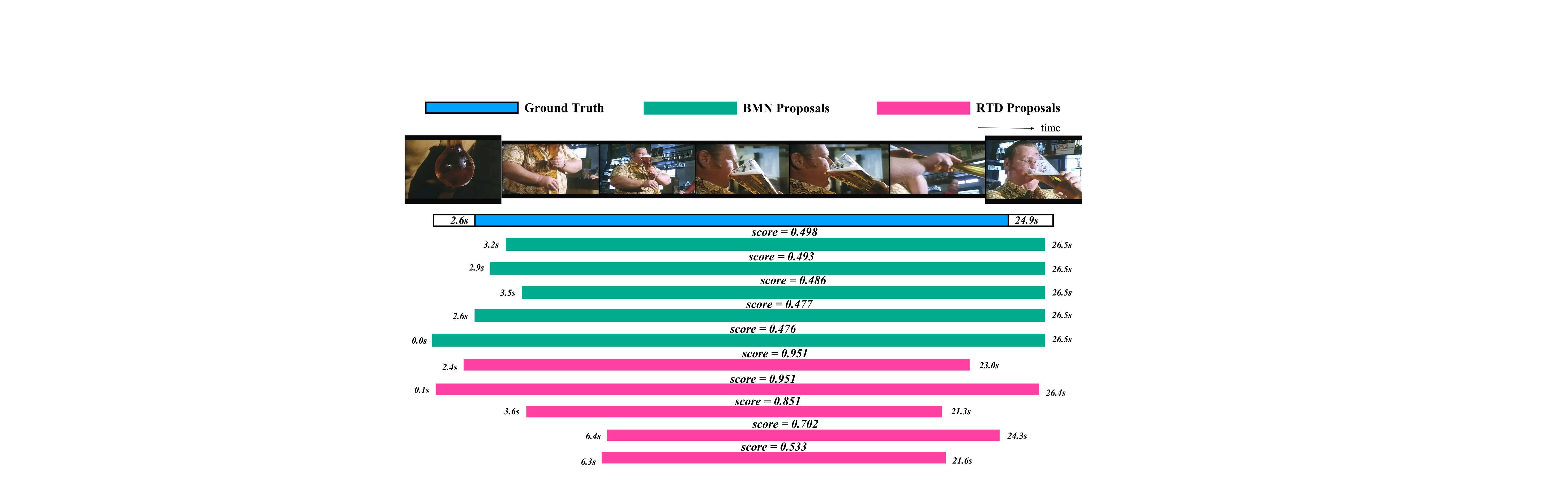}
\vspace{1mm}
\caption{Qualitative results of RTD-Net on ActivityNet-1.3. The proposals shown are the top-5 predictions for corresponding groundtruths based on the scoring scheme for each model.}
\label{fig:anet_qualitative}
\end{center}
\vspace{-2mm}
\end{figure}

\noindent\textbf{Qualitative results.} We visualize qualitative results in Figure~\ref{fig:anet_qualitative}. The top-5 predictions of BMN~\cite{DBLP:conf/iccv/LinLLDW19} share similar starting seconds and scores, and the same ending seconds. Bottom-up methods like BMN retrieve all proposals around locations with high boundary scores, while many of them are redundant and evaluated with similar confidence. If proposals around another groundtruth all have confidence over 0.9, the rankings of these proposals with confidence around 0.5 fall down, resulting in a low recall of this groundtruth. Therefore, heuristic NMS is introduced to address the above issues, which increases the inference time drastically. In contrast, a variation in localization appears in RTD predictions. Starting and ending locations of RTD proposals are varying from one another. More importantly, scores of RTD proposals are consistent with their rankings. Incomplete predictions are evaluated with lower scores, and ranked after those well-predicted proposals. As a result, RTD-Net is free of NMS module and has a much faster inference speed.

\section{Performance on HACS Segments}
\noindent\textbf{Dataset.} HACS Segments dataset~\cite{DBLP:conf/iccv/Zhao0TY19} contain 50,000 untrimmed videos and share the same 200 action categories with ActivityNet-1.3 dataset~\cite{DBLP:conf/cvpr/HeilbronEGN15}. To evaluate the quality of proposals, we calculate Average Recall with Average Number of proposals per video (AR@AN), and the Area under the AR vs AN curve (AUC) as metrics on HACS Segments dataset, which are the same as ActivityNet-1.3 dataset.

\noindent\textbf{Comparison with state-of-the-art methods.} We simply train RTD-Net on HACS Segments, with the same settings on ActivityNet-1.3. As Table~\ref{HACS} illustrates, RTD-Net achieves comparable results with only 100 queries per video. In contrast, BSN~\cite{DBLP:conf/eccv/LinZSWY18} predicts a large number of proposals and calculates evaluation metrics with top-100 of them. With top-100 proposals, BSN achieves a higher AR@100 than RTD-Net, while AUC of BSN and RTD-Net is the same. The comparison demonstrate RTD-Net achieves higher AR at small AN (e.g., AR@1), which indicates the efficiency of the direct action proposal generation mechanism. 

\begin{table}
\begin{center}
\caption{Comparison with other state-of-the-art proposal generation methods on validation set of HACS Segments in terms of AR@AN and AUC. Among them, only RTD-Net is free of NMS.}
\vspace{0mm}
\scalebox{0.85}{
\begin{tabular}{ccccccc}
\toprule[1pt]
Method & TAG+NMS~\cite{DBLP:conf/iccv/ZhaoXWWTL17} & BSN+SNMS~\cite{DBLP:conf/eccv/LinZSWY18} & RTD-Net \\ 
\hline
AR@1 (val) & - & - & \textbf{16.34} \\ 
AR@100 (val) & 55.88 & \textbf{63.62} & 61.11 \\ 
AUC (val) & 49.15 & \textbf{53.41} & \textbf{53.41}\\
\bottomrule[1pt]
\end{tabular}
}
\label{HACS}
\vspace{1mm}
\end{center}
\end{table}

\begin{figure}[t]
\begin{center}
\includegraphics[scale=0.5]{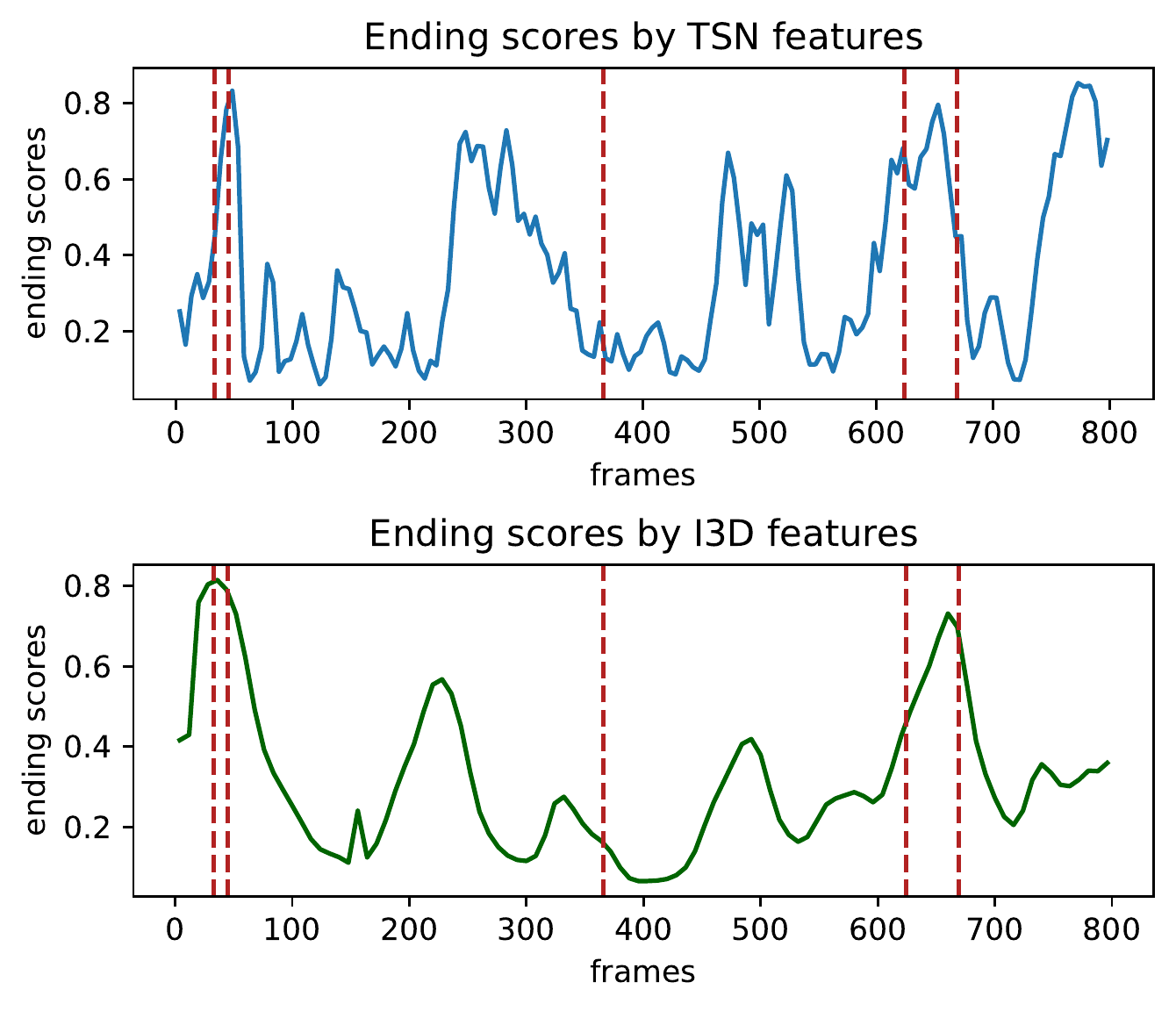}
\end{center}
\vspace{-2mm}
   \caption{Comparison of ending scores predicted by TSN and I3D feature extractors.}
\label{fig:boundary_comparison}
\vspace{-2mm}
\end{figure}

\section{Feature Encoding}
\noindent\textbf{Choices of feature extractors.}
There are two main types of feature extractors, one is 2D CNN (e.g., TSN~\cite{DBLP:conf/eccv/WangXW0LTG16}), the other captures temporal relations (e.g., I3D~\cite{DBLP:conf/cvpr/CarreiraZ17}). Bottom-up methods (e.g., BSN and BMN) first evaluate boundary confidence of all locations, and then explicitly match starting and ending points. With 2D CNN features that preserve local information better, bottom-up methods can achieve a higher recall of boundaries and better performance, which can be proved in the next section. Compared with 2D CNN features, 
I3D features have larger receptive fields and contain more temporal contexts. RTD-Net exploits self-attention blocks for proposal-proposal relations, and leverages encoder-decoder blocks to learn action-background differences. Therefore it can make full use of contextual information of I3D features and directly generate center locations and duration of proposals. 

\noindent\textbf{Comparison of boundary scores on different feature extractors.} According to the mechanism of the temporal evaluation module, temporal locations with boundary scores higher than a threshold or being with peak scores (namely their boundary scores $S_i$ are higher than their neighbors $S_{i-1}$ and $S_{i+1}$) are considered as candidates of action boundaries. Figure \ref{fig:boundary_comparison} displays the ending scores by TSN and I3D features, and groundtruth ending points are marked with vertical red dotted lines. We observe that TSN predictions covers every groundtruths with its local maximas but the first, achieving high recall of ending prediction. In contrast, the temporal evaluation module based on I3D features only captures the first groundtruth, resulting in a weaker recall. This might explains the performance drop of BSN and BMN with I3D feature input and gives solid support for our feature choice of the temporal evaluation module.

\noindent\textbf{Effect of feature modality.}
In Table \ref{modality}, we show the effect of feature modality on our framework by comparing the performance of RTD-Net under features from different modalities. We experiment with features from RGB, Optical flow and the fusion of both modalities. We find that Flow features outperforms RGB features by 1.5\% on AR@50, which indicates that motion information is more significant than appearance information in temporal action proposal generation. The fusion of both modalities here are in an early-fusion fashion, which requires both features concatenated in the beginning of the training and inference of the network. The early fusion features outperforms Flow features by 2.7\% on AR@50. 
\begin{table}[!h]
\begin{center}
\caption{Comparison of RGB and optical flow on THUMOS14, measured by AR@AN.}
\scalebox{0.8}{
\begin{tabular}{c|cccc}
\toprule[1pt]
Modality & @50 & @100 & @200 & @500 \\ 
\hline
RGB & 37.28 & 45.49 & 52.73 & 60.61 \\
Flow & {38.75} & {47.30} & {54.11} & {61.11} \\ 
Early Fusion &  \textbf{41.52} & \textbf{49.32} & \textbf{56.41} & \textbf{62.91} \\
\bottomrule[1pt]
\end{tabular}
}
\label{modality}
\end{center}
\vspace{-2mm}
\end{table}

{\small
\bibliographystyle{ieee_fullname}
\bibliography{egbib}
}

\end{document}